\pdfoutput=1

\documentclass[11pt]{article}
\usepackage{flushend}
\usepackage[preprint]{acl}
\usepackage{times}
\usepackage{latexsym}
\usepackage[T1]{fontenc}
\usepackage[utf8]{inputenc}
\usepackage{microtype}
\usepackage{inconsolata}
\usepackage{balance}
\usepackage{graphicx}
\usepackage{adjustbox}
\usepackage{booktabs}
\usepackage{subcaption}
\usepackage{multirow}
\usepackage{tikz}
\usepackage[noend]{algpseudocode}
\usepackage[inline]{enumitem}
\usepackage[colorinlistoftodos,prependcaption,textsize=tiny]{todonotes}
\usepackage{tikz}
\usepackage{enumitem}
\usepackage{ragged2e}
\usepackage[framemethod=TikZ]{mdframed}
\usepackage{amssymb}
\usepackage{xcolor}
\usepackage{times}
\usepackage{soul}
\usepackage{url}
\usepackage{hyperref}
\usepackage{anyfontsize}
\usepackage{graphicx}
\usepackage{amsmath}
\usepackage{amsthm}
\usepackage{booktabs}
\usepackage{algorithm}
\usepackage{subcaption}
\usepackage{xspace}
\usepackage{listings}
\usetikzlibrary{fit, positioning, decorations.pathreplacing, calc}
\newcommand{\systemname}{\textsc{MetaReflection}}

\title{\systemname: Learning Instructions for Language Agents using Past Reflections}
\author{
    \parbox{0.9\linewidth}{
        \centering{
            Priyanshu Gupta\thanks{ Equal Contribution}, \quad
            Shashank Kirtania\footnotemark[1], \quad 
            Ananya Singha\footnotemark[1]
           , \\
            Sumit Gulwani, \quad
            Arjun Radhakrishna,  \quad
            Sherry Shi, \quad 
            Gustavo Soares \\
            Microsoft \\
            {\tt \{priyansgupta, t-skirtania, ananyasingha, sumitg, arradha, shersh, gustavo.soares\}@microsoft.com}
        }
    }
}
\begin{document}
\maketitle

\newcommand\SelfReflect{\mathsf{SelfReflect}}
\newcommand\MetaReflect{\mathsf{MetaReflect}}
\newcommand\Predict{\mathsf{Predict}}
\newcommand\MergeDocs{\mathsf{MergeDocs}}

\newcommand\datapoint{\mathsf{dp}}
\newcommand\datapoints{\mathsf{Data}}
\newcommand\file{\mathsf{file}}
\newcommand\policy{\mathsf{policy}}
\newcommand\groundtruth{\mathsf{gt}}
\newcommand\ex{\mathsf{ex}}
\newcommand\examples{\mathsf{Ex}}
\newcommand\bool{\mathbb{B}}
\newcommand\docs{\mathbb{D}}
\newcommand\doc{\mathsf{doc}}
\newcommand\reflect{\mathsf{reflect}}
\newcommand\react{\textsc{ReAct}\xspace}
\newcommand\COT{\textsc{CoT}\xspace}
\newcommand\MetaReflection{\textsc{MetaReflection}\xspace}
\newcommand\protege{\textsc{ProTeGi}\xspace}
\newcommand\promptAgent{\textsc{PromptAgent}\xspace}
\newcommand\hotpot{\textsc{HotpotQA}\xspace}
\newcommand\gptfour{\textsc{GPT-4}\xspace}
\newcommand\gptturbo{\textsc{GPT-3.5-Turbo}\xspace}
\newcommand\iac{\textsc{IaC}\xspace}
\newcommand\biosses{\textsc{Biosses}\xspace}
\newcommand\bigbench{\textsc{BigBench}\xspace}
\newcommand\LLMInstr{\textsc{LLMInstruction}\xspace}
\newcommand\sdoc{\mathsf{sdoc}}
\newcommand\emptyDoc{\mathsf{doc}_\emptyset}

\newcommand\arsays[1]{{\color{purple}{AR: #1}}}
\newcommand\pgsays[1]{{\color{teal}{PG: #1}}}
\newcommand\sksays[1]{{\color{orange}{SK: #1}}}
\newcommand\gssays[1]{{\color{violet}{GS: #1}}}
\newcommand\sssays[1]{{\color{blue}{SS: #1}}}
\newcommand\assays[1]{{\color{brown}{AS: #1}}}
\newcommand\finsihed[1]{{\color{green}{AS: #1}}}

\newcommand\todos[1]{{\color{red}{todo: #1}}}
\newcommand\ignore[1]{}

\definecolor{myred}{RGB}{242,172,185}
\definecolor{mygreen}{RGB}{173,223,179}
\definecolor{myyellow}{RGB}{255, 255, 179}

\newcommand{\hlred}[1]{%
    {%
    \sethlcolor{myred}%
    \hl{#1}%
    }%
}
\newcommand{\hlgreen}[1]{%
    {%
    \sethlcolor{mygreen}%
    \hl{#1}%
    }%
}
\newcommand{\hlyellow}[1]{%
    {%
    \sethlcolor{myyellow}%
    \hl{#1}%
    }%
}

\begin{abstract}
The popularity of Large Language Models (LLMs) have unleashed a new age of \emph{Language Agents} for solving a diverse range of tasks. While contemporary frontier LLMs are capable enough to power reasonably good Language agents, the closed-API model makes it hard to improve in cases they perform sub-optimally. To address this, recent works have explored ways to improve their performance using techniques like self-reflection and prompt optimization. Unfortunately, techniques like self-reflection can be used only in an online setup, while contemporary prompt optimization techniques are designed and tested to work on simple tasks. To this end, we introduce \MetaReflection, a novel offline reinforcement learning technique that enhances the performance of Language Agents by augmenting a \emph{semantic memory} based on experiential learnings from past trials. We demonstrate the efficacy of \MetaReflection by evaluating across multiple domains, including complex logical reasoning, biomedical semantic similarity, open world question answering, and vulnerability threat detection, in Infrastructure-as-Code, spanning different agent designs. \MetaReflection boosts Language agents' performance by $4$ \% to $16.82$ \% over the raw \gptfour baseline and performs on par with existing state-of-the-art prompt optimization techniques while requiring fewer LLM calls. We release our experimental code at: \url{aka.ms/metareflection-code}.
\end{abstract}

\section{Introduction}


Large Language Models (LLMs), such as \gptfour~\cite{gpt4}, have gained significant popularity in recent years due to their ability to generate human-like text and solve complex tasks across various domains. To leverage these models, users typically craft prompts with instructions that are tailored to a specific task. Furthermore, many practical LLM applications setup complex multi-step systems with multiple LLM calls chained together \cite{langchain} or LLM calls with different prompts called in succession \cite{wu2023autogen}. Given their close resemblance to reinforcement learning agents, it has become useful to model these systems as Language Agents \cite{wu2023autogen}.


%
\ignore{
\begin{figure*}
\centering
\begin{tikzpicture}
\tikzstyle{box}=[
    draw,
    text width=3.5cm,
    minimum width=3cm,
    rounded corners=0.05cm,
    font=\fontsize{4}{5}\selectfont
]
\tikzstyle{labelbox}=[
    font=\fontsize{5}{6}\selectfont
]

\node[box] (task) {
Answer a question interleaving Thought, Action, and Observation
steps. Actions can be Search[\ldots], Lookup[\ldots], \ldots \\
\textbf{Question:} Did Jon Bon Jovi and Jeffrey Walker both form rock bands?
};
\node[labelbox, anchor=east, left=1mm of task.west] (task-label) {Task};
\node[labelbox, anchor=south, above=0mm of task.north] (test-label) {\textbf{Self-Reflection}};

\node[box, anchor=north, below=1mm of task.south] (trajectory) {
\textbf{Thought 1:} I need to search Jon Bon Jovi and Jeffrey Walker, find if they formed
rock bands, then find if both of them did.\\
\ldots\\
\textbf{Action 4:} Search[Jeffrey Walker]\\
\textbf{Observation 4:} Jeffrey Walker (born 10 July 1982) is an Australian
director and former actor, best known \ldots\\
\ldots\\
};
\node[labelbox, anchor=east, left=1mm of trajectory.west] (trajectory-label) {Trajectory};

\node[box, anchor=north, below=1mm of trajectory.south] (selfreflection) {
\textbf{Evaluation (feedback):} You attempted to answer the above question and
ran out of turns. Reflect on your mistakes \ldots \\
$\quad$ \\
\textbf{Self Reflection:} I made an assumption about the identity
of Jeffrey Walker based on the first search result. \ldots
In the future, I should consider the context of the question more carefully \ldots
For example, I could have searched for ``Jeffrey Walker musician" or ``Jeffrey Walker band" \ldots
};
\node[labelbox, anchor=east, left=1mm of selfreflection.west] (selfreflection-label) {Self-reflection};

\node[box, anchor=north, below=1mm of selfreflection.south] (trajectory-2) {
\ldots
};
\node[labelbox, anchor=east, left=1mm of trajectory-2.west] (trajectory-2-label) {Next trajectory};

\node[fit=(task) (trajectory) (selfreflection) (trajectory-2)] (run) {};

\draw [decorate,decoration={brace,amplitude=5pt,raise=0.5ex}]
  (selfreflection.north east) --  node (foo) {} (selfreflection.south east) {};

\node[box, right=1cm of task.north east, anchor=north west] (meta-reflection) {
Your job is to learn from experience of another agent and help it perform
better in the future.  Use the agent's reflections about its failures or success
and concretize them into instructions. \ldots\\
$\quad$\\
\textbf{Self-reflection 1:} I made an assumption about the identity of Jeffrey Walker \ldots\\
\textbf{Self-reflection 2:} \ldots
};
\node[labelbox, anchor=south, above=0mm of meta-reflection.north] (mr-label) {\textbf{Meta-Reflection}};

\node[box, anchor=north, below=1mm of meta-reflection.south] (instructions) {
\textbf{Instructions:}\\
1. When searching for a person, consider using their profession or nationality
as additional keywords to narrow down the results. \\
2. \ldots
};

\node[labelbox, right=3mm of selfreflection.south east, text width=1.3cm] (other-sr) {Self-reflections from other tasks};
\draw[->] (foo) ++ (2ex, 0cm) to[out=0, in=180, looseness=0.5] ($(meta-reflection.west) + (0cm, 0.3cm)$);
\draw[->, dashed] (other-sr.north) to[out=90, in=200, looseness=0.4] ($(meta-reflection.west) + (0cm, 0.0cm)$);


\node[box, right=2cm of meta-reflection.north east, anchor=north west, yshift=1cm] (test-task) {
Answer a question interleaving Thought, Action, and Observation
steps. Actions can be Search[\ldots], Lookup[\ldots], \ldots \\
\textbf{Question:} Which HGTV show does the american actor who hosted Monster
House appear in?
};
\node[labelbox, anchor=south, above=0mm of test-task.north] (test-label) {\textbf{Inference without meta-reflection}};

\node[box, anchor=north, below=1mm of test-task.south] (test-trajectory) {
\textbf{Thought 1:} \ldots\\
\textbf{Action 1:} Search[Monster House]\\
\ldots\\
\textbf{Action 4:} Search[Steve Watson]\\
\textbf{Obs. 4:} Steven Watson is an English football manager \ldots \\
\ldots\\
\textbf{Action 6:} Finish[There seems to be a discrepancy. Steve Watson is an
English football manager and does not appear in any HGTV shows.]
};

\node[box, below=1cm of test-trajectory.south, anchor=north] (test-task-2) {
Answer a question interleaving Thought, Action, and Observation
steps. Actions can be Search[\ldots], Lookup[\ldots], \ldots \\
$\quad$ \\
\textbf{Instructions:} \ldots When searching for a person, consider using their profession \ldots\\
\textbf{Question:} Which HGTV show does the american actor who hosted Monster
House appear in?
};
\node[labelbox, anchor=south, above=0mm of test-task-2.north] (test-label-2) {\textbf{Inference with meta-reflection}};

\node[box, anchor=north, below=1mm of test-task-2.south] (test-trajectory-2) {
\ldots\\
\textbf{Action 4:} Search[Steve Watson actor]\\
\textbf{Obs. 4:} Steven Watson is an American actor \ldots and currently serves
as the host of HGTV's Don't Sweat It.\\
\textbf{Action 6:} Finish[Don't Sweat It]
};

\draw[dotted]
    ($(test-task.north west) + (-1cm, 0.3cm)$)
    to
    node[labelbox, above, rotate=90] {Offline phase}
    node[labelbox, below, rotate=90] {Online phase}
    ($(test-trajectory-2.south west) + (-1cm, 0cm)$);

\draw[double]
    ($(test-trajectory.south west) + (-0.3cm,-0.3cm)$)
    to
    ($(test-trajectory.south east) + (+0.3cm,-0.3cm)$);

\draw[->] (instructions.east) to[out=0, in=180, looseness=0.7] (test-task-2.west) ;

\end{tikzpicture}
\caption{\MetaReflection in the HotPotQA. 
 reasoning\gssays{It would be good to show that the next trajectory in the training is a success. It would also be good to split the self-reflection box into 2: eval and self-reflection. Also the the feeback is not very clear for this example}}
\label{fig:workflow}
\end{figure*}
}
\begin{figure*}[!ht]
\centering
\begin{tikzpicture}
\tikzstyle{box}=[
    draw,
    text width=3.5cm,
    minimum width=3cm,
    rounded corners=0.05cm,
    font=\fontsize{5}{4}\selectfont
]
\tikzstyle{labelbox}=[
    font=\fontsize{6}{7}\selectfont
]

\node[box] (task) {
Answer a question interleaving Thought, Action, and Observation
steps. Actions can be Search[\ldots], Lookup[\ldots], \ldots \\
\textbf{Question:} Which member of FC Barcelona scored the only  goal in the 2010 FIFA World Cup final?
};
\node[labelbox, anchor=east, left=1mm of task.west] (task-label) {Task};
\node[labelbox, anchor=south, above=0mm of task.north] (test-label) {\textbf{Self-Reflection}};

\node[box, anchor=north, below=1mm of task.south] (trajectory) {
\textbf{Thought 1:} I need to search for the 2010 FIFA World Cup final and find
out who scored the only goal in that match.\\
\textbf{Action 1:} Search[2010 FIFA World Cup Final]\\
\textbf{Thought 2:} The paragraph does not mention who scored the goal...
\textbf{Action 2:} Lookup[goal]\\
\textbf{Obs. 2:} (Result 1/45) Sergio Ramos had a chance to score...\\
\textbf{Thought 3:} The observation still does not contain... I need to continue looking for ``goal"\\
\textbf{Action 3:} Lookup[goal]\\
\textbf{Obs. 3:} (Result 2/45) Robben missed an opportunity...\\
\ldots\\
};
\node[labelbox, anchor=east, left=1mm of trajectory.west] (trajectory-label) {Trajectory};

\node[box, anchor=north, below=1mm of trajectory.south] (selfreflection) {
\textbf{Evaluation (feedback):} You attempted to answer the above question and
ran out of turns. Reflect on your mistakes \ldots \\
$\quad$ \\
\textbf{Self Reflection:} In this trial, I failed to find the answer because I
was stuck in a loop of looking up goal, but the goal scorer was not found in the
first few results. This could be due to the fact that the information about the
goal might be located further down the page or in a different section...
};
\node[labelbox, anchor=east, left=1mm of selfreflection.west] (selfreflection-label) {Self-reflection};

\node[box, anchor=north, below=1mm of selfreflection.south] (trajectory-2) {
\ldots
};
\node[labelbox, anchor=east, left=1mm of trajectory-2.west] (trajectory-2-label) {Next trajectory};

\node[fit=(task) (trajectory) (selfreflection) (trajectory-2)] (run) {};

\draw [decorate,decoration={brace,amplitude=5pt,raise=0.5ex}]
  (selfreflection.north east) --  node (foo) {} (selfreflection.south east) {};

\node[box, right=1cm of task.north east, anchor=north west] (meta-reflection) {
Your job is to learn from experience of another agent and help it perform
better in the future.  Use the agent's reflections about its failures or success
and concretize them into instructions. \ldots\\
$\quad$\\
\textbf{Self-reflection 1:} In this trial, I was stuck in a loop...\\
\textbf{Self-reflection 2:} \ldots
};
\node[labelbox, anchor=south, above=0mm of meta-reflection.north] (mr-label) {\textbf{Meta-Reflection}};

\node[box, anchor=north, below=1mm of meta-reflection.south] (instructions) {
\textbf{Instructions:}\\
1. \hlyellow{If you're not finding the desired information or stuck in a loop of looking
up a keyword, consider changing the keyword and search strategy. The information
might be located further down the page.}\\
2. \ldots
};

\node[labelbox, right=3mm of selfreflection.south east, text width=1.3cm] (other-sr) {Self-reflections from other tasks};
\draw[->] (foo) ++ (2ex, 0cm) to[out=0, in=180, looseness=0.5] ($(meta-reflection.west) + (0cm, 0.3cm)$);
\draw[->, dashed] (other-sr.north) to[out=90, in=200, looseness=0.4] ($(meta-reflection.west) + (0cm, 0.0cm)$);


\node[box, right=2cm of meta-reflection.north east, anchor=north west, yshift=1cm] (test-task) {
Answer a question interleaving Thought, Action, and Observation
steps. Actions can be Search[\ldots], Lookup[\ldots], \ldots \\
\textbf{Question:}  Who was also an airline pilot, Billy Gibbons or Bruce Dickinson?
};
\node[labelbox, anchor=south, above=0mm of test-task.north] (test-label) {\textbf{Inference without meta-reflection}};

\node[box, anchor=north, below=1mm of test-task.south] (test-trajectory) {
\ldots \\
\textbf{Thought 3:} \ldots\\
\textbf{Action 3:} Search[Bruce Dickinson]\\
\textbf{Obs. 3:} Paul Bruce Dickinson is an English singer, best known...\\
\textbf{Thought 4:} There is no mention of Bruce Dickinson being an airline pilot in this paragraph...\\
\textbf{Action 4:} Lookup[airline pilot]\\
\textbf{Obs. 4:} No more results\\
\ldots\\
\textbf{Action 6:} Search[Bruce Dickinson (pilot)]\\
\hlred{\textbf{Obs. 6:} Out of turns}
};

\node[box, below=1cm of test-trajectory.south, anchor=north] (test-task-2) {
Answer a question interleaving Thought, Action, and Observation
steps. Actions can be Search[\ldots], Lookup[\ldots], \ldots \\
$\quad$ \\
\textbf{Instructions:} \hlyellow{\ldots If you are not finding the desired information \ldots}\\
\textbf{Question:} Who was also an airline pilot, Billy Gibbons or Bruce Dickinson?
};
\node[labelbox, anchor=south, above=0mm of test-task-2.north] (test-label-2) {\textbf{Inference with meta-reflection}};

\node[box, anchor=north, below=1mm of test-task-2.south] (test-trajectory-2) {
\ldots\\
\textbf{Action 4:} Lookup[airline pilot]\\
\textbf{Obs 4:} No more results\\
\textbf{Thought 5:} The information about Bruce Dickinson being an airline pilot might be located further down the page.
I need to look up the related term ``pilot"\\
\textbf{Action 5:} Lookup[pilot]\\
\textbf{Obs. 5:} (Result 1/10) He undertook a career as a commercial pilot ...\\
...
\hlgreen{\textbf{Action 6:} Finish[Bruce Dickinson}]
};

\draw[dotted]
    ($(test-task.north west) + (-1cm, 0.3cm)$)
    to
    node[labelbox, above, rotate=90] {Offline phase}
    node[labelbox, below, rotate=90] {Online phase}
    ($(test-trajectory-2.south west) + (-1cm, 0cm)$);

\draw[double]
    ($(test-trajectory.south west) + (-0.3cm,-0.3cm)$)
    to
    ($(test-trajectory.south east) + (+0.3cm,-0.3cm)$);

\draw[->] (instructions.east) to[out=0, in=180, looseness=0.7] (test-task-2.west) ;

\end{tikzpicture}
\caption{\MetaReflection in the \hotpot domain with a \react agent 
}
\label{fig:workflow}
\end{figure*}
%



Building up on these insights, recent works~\cite{shinn2023reflexion, madaan2023selfrefine} have showed that the performance of such language
agents can be improved using \emph{verbal reinforcement learning} across
multiple conversational turns, by providing feedback at the end of a failing trajectory and
asking the agent to reflect on its mistakes. The reflective text is then stored as episodic
memory to improve future trajectories on the same task.
%

Parallel to this, there has been recent advancements in developing prompt optimization techniques \cite{pryzant-etal-2023-automatic, wang2023promptagent}. These techniques typically start with a seed human prompt and a train dataset of Input-Output pairs. Leveraging error feedback over failing cases, they deploy various search strategies to find prompt variants that can give the optimal results on train set.

While prompt optimization techniques exist for single LLM call setups, they are not designed for complex multi-step language agents. Here, techniques like self-reflection can help improve a language agent's performance significantly. However, these are online reinforcement processes that depend on the
availability of performing multiple turns with a feedback mechanism on the same task instance and aren't applicable to similar new tasks. 


Motivated by the \emph{semantic memory} observed in humans \cite{mcrae201314, Weng_2023_blog, Paischer2023SemanticHA} we introduce \MetaReflection, an offline reinforcement learning technique that works by augmenting a \emph{semantic memory} to represent experiential learnings from trials in an offline training phase. At a high level, \MetaReflection works by simulating various trials with samples from a training dataset, gather self-reflections from failing trials and iteratively builds an experiential \emph{semantic memory} by generalizing the self-reflections into `meta-reflections' that takes the form of instructions to the language agent.

We evaluate the \emph{\MetaReflection} technique across multiple scenarios: 3 complex logical reasoning datasets \cite{suzgun2022challenging}, a biomedical semantic similarity dataset \cite{souganciouglu2017biosses}, an open world question answering \cite{yang2018hotpotqa}, and a new vulnerability threat detection in a new Infrastructure-as-Code (\iac) dataset.

The \iac vulnerability detection dataset, in particular, is a new work that we are introducing with this paper, entailing detection of 10 unique common security vulnerabilities found in Terraform \cite{terraform} cloud infrastructure configuration files. While static analysis tools like Terrascan \cite{terrascan} exist, they are not able to detect semantic nuances beyond the standard hard-coded detection patterns. Being a low-resource language,  vanilla \gptfour struggles to effectively detect the vulnerabilities. This makes it a particularly interesting dataset for domain-specific reasoning.  

\MetaReflection achieves $4$\% to $16.82$\% higher accuracy over raw \gptfour  baseline. For setups, where they are directly comparable, \MetaReflection performs on par or better than state of the art prompt optimization techniques, while requiring fewer LLM calls for learning. 
We also demonstrate how \MetaReflection can be applied to multi-step agents for offline learning.

To summarize, we make the following contributions:
\begin{itemize}
    \item We present \MetaReflection, a technique for learning \textit{semantic memory} for language agents using offline training simulations. To the best of our knowledge, this is the first approach towards learning semantic memory for generic language agents.
    \item We introduce a new dataset for vulnerability threat detection in \iac. The dataset comprises $416$ challenging vulnerability detection tasks. 
    \item We conducted an extensive evaluation of the \emph{\MetaReflection} technique across multiple distinct domains: vulnerability threat detection, causal reasoning, domain-specific semantic similarity, and open-domain question answering, demonstrating strong improvements in agent performances. (Section~\ref{sec:eval}).

\end{itemize}
\section{\MetaReflection}
\label{sec:technique}

\begin{algorithm}
\small
    \begin{algorithmic}[1]
        \Require
            Train data $D_T$, Val. data $D_V$,
            Agent $\mathsf{A}$
        \Require
            Environment $\mathsf{Env}$
        \Ensure Meta-reflection instructions $\mathsf{Instr}$
        \State $\mathsf{Instr} \gets \bot$
        \ForAll { $\mathsf{b} \in \mathsf{Batch}(D_T)$}
        \State ${\mathsf{v} \gets \mathsf{RandomSample}(D_V)}$
        \ForAll{$\mathsf{Trials} \in \{0,1 \ldots \mathsf{MaxTrials}\} $}
        
        \label{line:iterative-refinement}\Comment{Iterative Refinement}
        \State Initialize $\mathsf{A}$ with $\mathsf{Instr}$
            \label{line:seed}
        \State $\mathsf{trajectories} \gets \{\mathsf{A.Run(\textit{i})} \mid i \in \mathsf{b}~~\} $
         \State $\mathsf{fT} \gets \mathsf{Env.GetFailingTrajectories(trajectories)} $
         \If{$\mathsf{fT} == \{ \}$}
         \State $\mathsf{break}$ \Comment{Early Stopping}
         \EndIf
        \State $\mathsf{refl} \gets \{~~\mathsf{A}.\mathsf{SelfReflect}(t)  \mid t \in \mathsf{fT}~~\}$
        \label{line:self-reflect}
        \State $\mathsf{Instr}^* \gets \mathsf{A.MetaReflect}(\mathsf{Instr}, \mathsf{b}, \mathsf{refl})$
            \label{line:meta-reflection}
        
        \If { $\mathsf{A.ShowsImprovement}(\mathsf{Instr}^*, \mathsf{Instr}, \mathsf{b},\mathsf{v}, \mathsf{Env})$} 
            \label{line:validation}\Comment{Validation}
        \State $\mathsf{Instr} \gets \mathsf{Instr}^*$
        \label{line:back-track}\Comment{Else Backtrack}
        \EndIf
        \EndFor
        \EndFor
        \State \Return $\mathsf{Instr}$
    \end{algorithmic}
    \caption{\MetaReflection}
    \label{algo:meta-reflection-with-batch-chaining}
\end{algorithm}

We consider a setup of an LLM Agent $\mathsf{A}$ interacting with an environment $\mathsf{Env}$ to solve a task $\mathsf{T}$, characterized by the start state $\mathsf{S_0}$.
At time step $t$, the Agent $\mathsf{A}$ receives an observation $\mathsf{o_t}$ at state $\mathsf{S_{t}}$ and samples an action $\mathsf{a_t}$ using a policy  $\pi(\mathsf{a}_t|\mathsf{T}_{t}, \mathsf{P}_t)$ where $\mathsf{T}_{t} = \{\mathsf{o_0},\mathsf{a_0},\mathsf{o_1} \ldots,\mathsf{a_t}, \mathsf{o_t}\}$, represents the trajectory of the agent and the prompt $\mathsf{P_t}$ is the prompt that drives the action selection at time step $t$. Eventually, the agent reaches a state $S_n$ where it receives a reward from the agent $\mathsf{r= R(S_n)}$. This state marks the  correct/incorrect completion of the task  $\mathsf{T}$ by the language agent. 

\MetaReflection{} aims to improve the performance of such language agents by augmenting them with a \emph{semantic memory} \cite{mcrae201314} in the form of rule based instructions $\mathbf{Instr}$ to store the experiential learnings, replacing $\pi(\mathsf{a_t}|\mathsf{T}_t, \mathsf{P})$ 
with $\pi(\mathsf{a_t}|\mathsf{T}_t, \mathsf{P \bigoplus Instr})$. Algorithm~\ref{algo:meta-reflection-with-batch-chaining} shows the
outline of the \MetaReflection process.
At its core, the algorithm works by starting with an empty set of instructions
and iteratively improving the instructions using small training batches.

For instance, if the agent $\mathsf{A_{react}}$ is a \react- style agent working on a Question Answering Task over Wikipedia, at state $S_0$, the agent will be provided by a question $Q$. Now, the agent can choose to take a $\mathsf{search}$ action, upon which it will receive an observation $o$. The agent can then either chose to pick further $\mathsf{search}$ or $\mathsf{lookup}$ actions, or pick an $\mathsf{answer}$ action to return an answer. This action will mark a transition to the state $\mathsf{S_n}$ and a binary reward can be generated depending on the correctness of the answer. An even simpler example can be an agent $\mathsf{A_s}$ which given a Multiple-Choice Question, picks an answer and receives a positive reward if the answer is correct.


\paragraph{Agent, Environment and Self-reflection method} \label{para:Agents}
\MetaReflection procedure
uses the following components at its core:

\begin{enumerate*}[label={(\alph*)}]

\item an agent $\mathsf{A}$ (i.e., an RL actor) that is based on a language model,
\item an environment $\mathsf{Env}$ that generates rewards for specific actions, given a
  agent's trajectory, and
\item A $\mathsf{SelfReflect}$ method on the lines of \cite{shinn2023reflexion} that produces a verbal reinforcement given an agent's
  trajectory (line~\ref{line:self-reflect}, Algorithm~\ref{algo:meta-reflection-with-batch-chaining})
\end{enumerate*} This is in contrast to prior state of the art prompt optimization techniques \cite{wang2023promptagent, pryzant-etal-2023-automatic}, where optimization relies solely on the final action of the agent and cannot be applied on the intermediate un-observed states of the actor. 
Additionally, we adapt the agent $\mathsf{A}$ to be parameterized by a set of
instructions in addition to the standard task description and agent behaviour prompt.
%
In our implementation and experiments, we utilize either a single-step agent (a vanilla single-shot Language Model agent) or a multi-step agent based on \react~\cite{yao2023react} and \COT \cite{wei2023chainofthought}. While \MetaReflection makes no distinction between the two, we make this distinction to ease comparison with other prompt optimization techniques, which only work in the single-step agent configuration. More information on agent configurations can be found in Section~\ref{subsec:agent-config}.
%
%

\paragraph{The $\mathsf{MetaReflect}$ step}

The $\mathsf{MetaReflect} $ method is designed to take as input a prior set of
instructions $\mathsf{Instr}$, a set of self-reflections $\mathsf{refl}$,
the training data $\mathsf{D_T}$ and a validation data $\mathsf{D_V}$ and will produce an updated set of instructions
$\mathsf{Instr^*}$ as denoted in line~\ref{line:meta-reflection} of Algorithm~\ref{algo:meta-reflection-with-batch-chaining}.

For the $\mathsf{MetaReflect}$ method, we use a standard language model with a prompt
that instructs the LLM to observe the reflections, the training data, and
produce a new non-case specific instruction style memory.
Further, the prior memory is also passed as input so that the output is a
generalization of the prior learnings. 
%
In our implementation, this meta-reflection and generalization are done in the
same prompt for efficiency.
%
%
Alternatively, new instructions can be generated first and then combined with existing ones.

We specify that the instructions need to take the form of a list.
Hence, the meta-reflection step in line~\ref{line:meta-reflection} typically either
\begin{enumerate*}[label=(\alph*)]
\item updates the list by adding a new item, or
\item combines one or more previous items with learnings from the
  self-reflections to produce a shorter list. 
\end{enumerate*}
For example, one meta-reflection instruction learned during our \hotpot
experiments suggested including the profession when searching for a person to
narrow down results.
In a subsequent batch, the self-reflection step produces a reflection that
mentions adding search terms like release date when searching for
movies.
The $\mathsf{MetaReflect}$ step may combine the previous instructions with the current
self-reflections either by appending a new item to the list clarifying the
strategy to search for movies, or may generalize the previous item to something
like ``\textit{When searching for specific entities, use additional contextual
information to augment the primary search terms with secondary keywords
corresponding to the characteristics of the entity}".

\paragraph{Validation and Backtracking}
In each iteration, after $\mathsf{MetaReflect}$, we validate the quality of
the new instructions.
Due to the sparse reward signals leading to poor self-reflections or
over-generalization of the meta-reflection instructions, we may end up with
instructions that are of a poorer quality than the prior instructions.
The poorer instructions may also be due to general capricious, unpredictable
nature of large language models.
Therefore, we validate the new instructions by testing them on training data and a random sample of the validation set to
ensure that they perform better than the prior instructions as depicted in line~\ref{line:validation} of Algorithm~\ref{algo:meta-reflection-with-batch-chaining}.  
Ideally, we would do this validation over the full validation dataset.
However, in our case,
we only validate on the current batch to balance
quality of instructions and efficiency.
In cases where the updated instructions perform poorly compared to the prior ones, we \emph{backtrack} to prior instructions as depicted in line~\ref{line:back-track}.

As an example, in the previous paragraph the meta-reflection step replaced
the specific instruction on how to search for persons with a more general
instruction on how to search for entities.
However, it is possible that these general instructions are too vague
(especially for smaller, less capable models) and the client agent is not able
to apply them correctly to the case of searching for persons.

\paragraph{Iterative Refinement}
%
%
Motivated by the self refining behaviour of Language models as demonstrated in \cite{shinn2023reflexion}, we use multiple attempts at meta-reflection for each
batch (line~\ref{line:iterative-refinement} through line~\ref{line:back-track}) until we see no failures in the current batch or exhaust a maximum number of trials (set to $3$ for all our experiments). Similar to how self-reflections  help optimize Language agents' trajectories towards achieving a task, the feedback on failed trajectories from the $\mathsf{Instr^*}$ act as implicit \emph{verbal reinforcements} in the meta-reflection process. These \emph{verbal reinforcements} can then be used by the meta-reflection step to guide the instruction search. This trajectory driven iterative refinement strategy reduces the chances of repeating mistakes, such as proposing incorrect refinements that were already tried in the past, during the instruction refinement process.

In the single step agent setups where they are directly comparable, we observe that this iterative refinement strategy leads us to similar performance as state of the art prompt optimization techniques, with lesser number of LLM calls. 



\ignore{
High level plan:
\begin{itemize}
    \item Talk about Terraform \iac
    \item Why Security vulnerability is a problem - degree of issues (why is it a hard problem)
    \item Low-resource language, so poorer performance.
    \item Data creation:
    \begin{itemize}
        \item Identified data from github repos with \iac code.
        \item Manually annotated 202 files for 10 policies. 
        \item (Optional) [Process for data creation] Semantic search + GPT4 step to do coarse filtering of policies (pick top 10 (confirm with sherry) most relevant policies). Ran a static analysis tool - Terrascan for further validation. Validated clashes with Self-Reflection. If still clash, manually validate. 
    \end{itemize}
    \item Plots for distribution of true/test/true positives/false positives.
    \item Results:
        \begin{itemize}
            \item POlicywise Train/test splits numbers. Data created keeping the following in mind:
                \begin{itemize}
                    \item Train : Test :: 40:60 
                    \item The rarer gt label is no more than twice the more common label (justify)
                    \item (Optional) The FN, FP are equally split and train and test and are never sub sampled. This ensures that the failing GPT-4 cases are well represented in the dataset.
                \end{itemize}
            \item Policy-wise accuracy and recall numbers across batch sizes and baseline
            \item Some example of learned instructions for policy.
            \item 1 example of succeeding case. 
        \end{itemize}
\end{itemize}
}

\section{Experimental Setup}
\label{sec:eval}

\subsection{Datasets}
We evaluate \MetaReflection on datasets from different domains like vulnerability threat detection (\iac), question answering (\hotpot), Complex Reasoning (\bigbench), Biomedical Semantic Similarity (\biosses).

\paragraph{Vulnerability Threat Detection (\iac)} 
Infrastructure-as-Code (\iac) is a popular method of configuring cloud infrastructures,
on platforms such as Azure and AWS, using a configuration coding language. 
Here, we focus on Terraform, a leading \iac platform by
Hashicorp~\cite{terraform}, as well as Azure, the cloud computing platform by Microsoft, which comes as a reusable configuration component.
Such Cloud infrastructures are prone to security vulnerabilities such as open ports and
exposed administrator accounts~\cite{terrascan}.
Vulnerability detection via static analysis such as Terrascan~\cite{terrascan} of \iac files
is a hard problem due to the expressivity and complexity of the configuration language and the diversity of the resources being
handled across multiple infrastructure providers (e.g., Amazon AWS and Microsoft
Azure). 

This opens up the possibility of using an LLM to perform vulnerability detection which entails checking if a given Terraform module violates
a given Terrascan policy.
To evaluate the efficacy of \MetaReflection on vulnerability detection task we collected  $202$ Terraform modules by mining GitHub repositories and post processing it further to achieve 186 data points which is then split $40:60$ for train and test respectively. You can find more information on \iac, Terraform, Terrascan and data collection in Appendix~\ref{sec:iac-appendix}.

\paragraph{Complex Reasoning (\bigbench)}
Big-Bench Hard (BBH) \cite{suzgun2022challenging} consists of a subset of particularly challenging tasks from BIG-Bench \cite{srivastava2022beyond} that contains challenging reasoning  questions. In this work, we pick 3 datasets from BBH - \begin{enumerate*}
    \item \textbf{Causal Judgement}
    \item \textbf{Temporal Sequence}
    \item \textbf{Epistemic Reasoning}
    \end{enumerate*}
    and follow the exact same  test and train distribution used in \cite{wang2023promptagent}. Table~\ref{tab:train-test-set} shows the test and the train distribution for the benchmarks. 

\begin{table}[]
    \centering
    \small
    \begin{tabular}{l ccc}
    \toprule
    Dataset & Train set & Test set\\
    \midrule
    \biosses & 60 & 40 \\

    \textsc{Casual Judgement} & 90& 100 \\
    \textsc{Epistemic Reasoning} & 500& 500 \\
    \textsc{Temporal Sequence} & 300& 500 \\
    
    \textsc{\iac} Vulnerability Detection & 166 & 250\\
    \textsc{\hotpot} & 50 & 80\\
    \bottomrule
    \end{tabular}
    \caption{Count of train and test set distribution across benchmarks}
    \label{tab:train-test-set}
\end{table}
\paragraph{Biomedical Semantic Similarity (\biosses)} is a biomedical sentence similarly dataset \cite{souganciouglu2017biosses}. Each instance in the dataset comprises of two sentences which are to be compared. As prior work \cite{wang2023promptagent}, we model the problem as a classification task between labels `similar', `non-similar' or `somewhat similar' and utilize the exact same test and train data splits.


\paragraph{Question Answering (\hotpot)} 
The dataset \cite{yang2018hotpotqa} is a Question-Answering dataset  consisting of $113k$ question-and-answer pairs over Wikipedia. A typical system working over this dataset first performs a retrieval over wikipedia and the reasons over the retrievals to come up with an answer.
Besides the retrieve and reason setup, the dataset also comes in 2 resoning only settings: \begin{enumerate*}
    \item \textbf{GT}: where each Question is accompanied with the most relevant  supporting \emph{ground truth} documents and 
    \item \textbf{Distractor} where the Question is accompanied with the ground truth documents, alongside some distracting documents making context analysis and interpretation more challenging. 
\end{enumerate*} 
Given the large scale of the dataset, we adversarially select (Appendix \ref{subsec:appendix-hotpotqa-data-sampling}) test samples from the test split of the dataset to ensure good sample diversity.To perform adversarial sampling, we identify samples where the Agent $\mathbf{A}$ consistently fails and conduct up to three self-reflection trials to correct the response. If the Agent $\mathbf{A}$ still fails, we discard these samples. This method ensures we gather challenging examples while filtering out noisy ones, and we sample $40$ and $80$ examples for the \react train and test sets, and $50$ and $80$ for \COT settings, respectively.

\subsection{Baselines}
In absence of techniques for direct comparison, we compare \MetaReflection against strong contemporary prompt optimization baselines \begin{enumerate*}
    \item \protege and 
    \item\promptAgent
\end{enumerate*} across multiple single step agent settings.  

\paragraph{\protege} \cite{pryzant-etal-2023-automatic} performs prompt optimization by leveraging batch-wise error feedback as textual gradients and use them to generate multiple variations of the current prompt. Notably, these textual gradients are prompt update `guidelines' generated by looking at the failing error output. They then deploy beam search to find the best prompt by iteratively evaluating and generating more prompt candidates.

\paragraph{\promptAgent}\cite{wang2023promptagent}
Similar to \protege, \promptAgent also leverages error feedback to generate prompt candidates. They further optimize prompt search using a principled Monte-Carlo Tree Search and identify high-reward paths to find the best prompts.


\subsection{Agent Configurations}
\label{subsec:agent-config}
The generic Language Agent $\mathsf{A}$ defined in Section \ref{para:Agents} can operate upon multiple time steps before reaching a response state $\mathbf{S_n}$. While, the \MetaReflection algorithm is broadly applicable to all such agents, for the purpose of simplification of comparison, we distinguish our agent setups in 2 categories:\begin{enumerate*}
    \item \emph{Single-Step}  and
    \item \emph{Multi-Step} Agents.
\end{enumerate*}

\paragraph{Single-Step Agent}
We call an Agent $\mathsf{A}$ a Single-Step Agent if it takes exactly 1 time step starting from the initial state $S_0$ to reach an answer state $S_1$, without traversing intermediary belief states. In practice, what it means is that the agent is characterized by a single zero-shot prompt that always elicits the agent to generate a response solving the input task. We define such single step agents for the \iac, \biosses and the \bigbench datasets.    


\paragraph{Multi-Step Agent}
\label{para:Multi-Step}
We define Multi-Step Agents to be  agents with atleast one intermediate belief state. In our evaluations we consider two well established multi-step agent style \COT \cite{wei2022chain} and \react
\cite{yao2022react}. The \COT setup entails exactly 2 time steps - the first transition generates the `thought' and the next transition generates the `answer'. The \react setup, however, can entail multiple time steps with `actions' for context gathering, before generating an `answer' action. For the purpose of our evaluation, similar to \cite{shinn2023reflexion}, we use the different \hotpot datasets to setup \COT-(Distractor) and \COT-(GT) for the Distractor and GT settings respectively and a \react agent with $\mathsf{search}$ and $\mathsf{lookup}$ actions on the wikipedia corpus for the retrieval and reasoning setting. 

\subsection{Other Experimental Configurations}
We use \gptfour-32k chat model as an LLM throughout the experiments, with a $temperature = 0$. For \MetaReflection, we set the $\mathsf{batch}$ size to be 4 and $\mathsf{maxRetries}$ to be 3, and the random sample size for the validation set to be 5. To account for statistical variations, all the experiments are run for 3 runs and we report the averages and standard deviation for each of them.

\section{Results}
\label{sec: Results and Analysis}
\begin{table*}[t]
\centering
\resizebox{\textwidth}{!}{%
\begin{tabular}{@{}lclcllcllcll@{}}
\toprule
\multicolumn{1}{c}{Dataset} & \multicolumn{2}{c}{\gptfour} & \multicolumn{3}{c}{\protege} & \multicolumn{3}{c}{\promptAgent} & \multicolumn{3}{c}{\MetaReflection} \\ 
\cmidrule(lr){2-3} \cmidrule(lr){4-6} \cmidrule(lr){7-9} \cmidrule(lr){10-12} 
                             & Acc   & $\sigma$ & Acc   & $\sigma$ & \# calls & Acc         & $\sigma$ & \# calls & Acc            & $\sigma$ & \# calls       \\ \midrule
\biosses                     & 70.00 & 3.17     & 80.75 & 0.71     & 6478     & 80.00          & 1.34     & 503      & \textbf{84.00}    & 0.57     & $\textbf{475}$ \\ \midrule
\textsc Causal Judgement          & 74.00    & 1.15     & 72.20 & 3.67     & 9725     & 73.40       & 3.19     & 351      & \textbf{77.00}    & 2.00     & \textbf{313}   \\
\textsc Epistemic Reasoning             & 70.80  & 1.03     & 85.40 & 2.40     & 16640    & \textbf{91.00} & 0.83     & 2143     & 88.40           & 1.02     & \textbf{1954}  \\
\textsc Temporal Sequence     & 98.00  & 0.57     & 97.50  & 0.27     & 13824    & 98.30        & 1.11     & 2017     & \textbf{99.10}  & 0.63     & \textbf{1491}  \\ \midrule
\iac Vulnerability Detection & 73.36 & 0.01     & 72.64 & 4.76     & 8148     & 84.17       & 6.91     & 354      & \textbf{90.18} & 0.28     & \textbf{166}   \\ \bottomrule
\end{tabular}%
}
\caption{Comparison of \MetaReflection with other baselines in a single-step agent set-up.\texttt{\# calls} refers to average numbers of LLM calls required to generate optimal prompt}
\label{tab:RQ1--single-agent-baseline-comparision}
\end{table*}

\begin{table*}[]
\resizebox{\textwidth}{!}{%
\begin{tabular}{@{}lccccccc@{}}
\toprule
\multicolumn{1}{c}{\multirow{2}{*}{Policy}} &
  \multicolumn{3}{c}{Data statistics} &
  \multirow{2}{*}{\gptfour} &
  \multirow{2}{*}{\protege} &
  \multirow{2}{*}{\promptAgent} &
  \multirow{2}{*}{\MetaReflection} \\
\multicolumn{1}{c}{}                & Files & Violating & Compliant &    &       &                &                \\ \midrule
networkPolicyEnabled                & 23    & 10        & 13        & 48 & 62.18 & 71.06          & \textbf{83.34} \\
kubeDashboardDisabled               & 29    & 3         & 26        & 89 & 86.22 & 90.80 & \textbf{94.63}          \\
reme\_keyVaultAuditLoggingEnabled   & 44    & 10        & 34        & 55 & 77.97 & 82.71          & \textbf{98.18} \\
reme\_containerRegistryResourceLock & 35    & 9         & 26        & 53 & 68.37 & 90.22 & \textbf{96.55}          \\
reme\_appGatewayWAFEnabled          & 59    & 36        & 23        & 72 & 68.06 & 88.70           & \textbf{92.30} \\
networkPort22ExposedToInternetAz    & 27    & 4         & 23        & 94 & 87.84 & 89.02          & \textbf{96.29} \\
networkPort22ExposedToPublicAz      & 35    & 6         & 29        & 94 & 85.22 & 80.60           & \textbf{94.11} \\
reme\_noSecurityGroupAssociated     & 90    & 60        & 30        & 70 & 67.20  & 82.60           & \textbf{87.58} \\
reme\_checkStorageContainerAccess   & 19    & 5         & 14        & 91 & 88.62 & 88.05          & \textbf{100.00}   \\
reme\_resourceGroupLock             & 55    & 43        & 12        & 81 & 80.33 & 77.98          & \textbf{82.90}  \\ \bottomrule
\end{tabular}%
}
\caption{Performance of \MetaReflection and other baselines on \iac tasks using singe-step agent}
\label{tab:RQ-1--Iac-per-policy-single-agent-baseline-comparision}
\end{table*}




\subsection{Comparison with the baselines}
Table~\ref{tab:RQ1--single-agent-baseline-comparision} and  Table~\ref{tab:RQ-1--Iac-per-policy-single-agent-baseline-comparision} compares \MetaReflection with baseline \gptfour and various prompt optimization baselines over multiple datasets in a single-step agent setting. Further, Table~\ref{tab:RQ1--single-agent-baseline-comparision} also presents the average number of LLM calls it took by different systems to generate the optimal prompt. 

We observe that \MetaReflection fairs competitively with the other techniques, outperforming  \gptfour and \protege over all the benchmarks. Compared to \promptAgent we observe better performance in all the datasets except for \textsc{Epistemic Reasoning} where we see a slight regression. Notably, in all the configurations, \MetaReflection required the least number of LLM calls for training (include any intermediate inference calls), with \protege and \promptAgent requiring upto $17$ and $1.5$ times more LLM calls.
In general, we observe that even with a lower number of required LLM calls, \MetaReflection performs at par with \promptAgent for generic causal reasoning  tasks, while outperforming it on highly contextual domain-specific settings like \iac and \biosses.

\begin{figure}[h]
  \centering
  \includegraphics[width=0.5\textwidth]{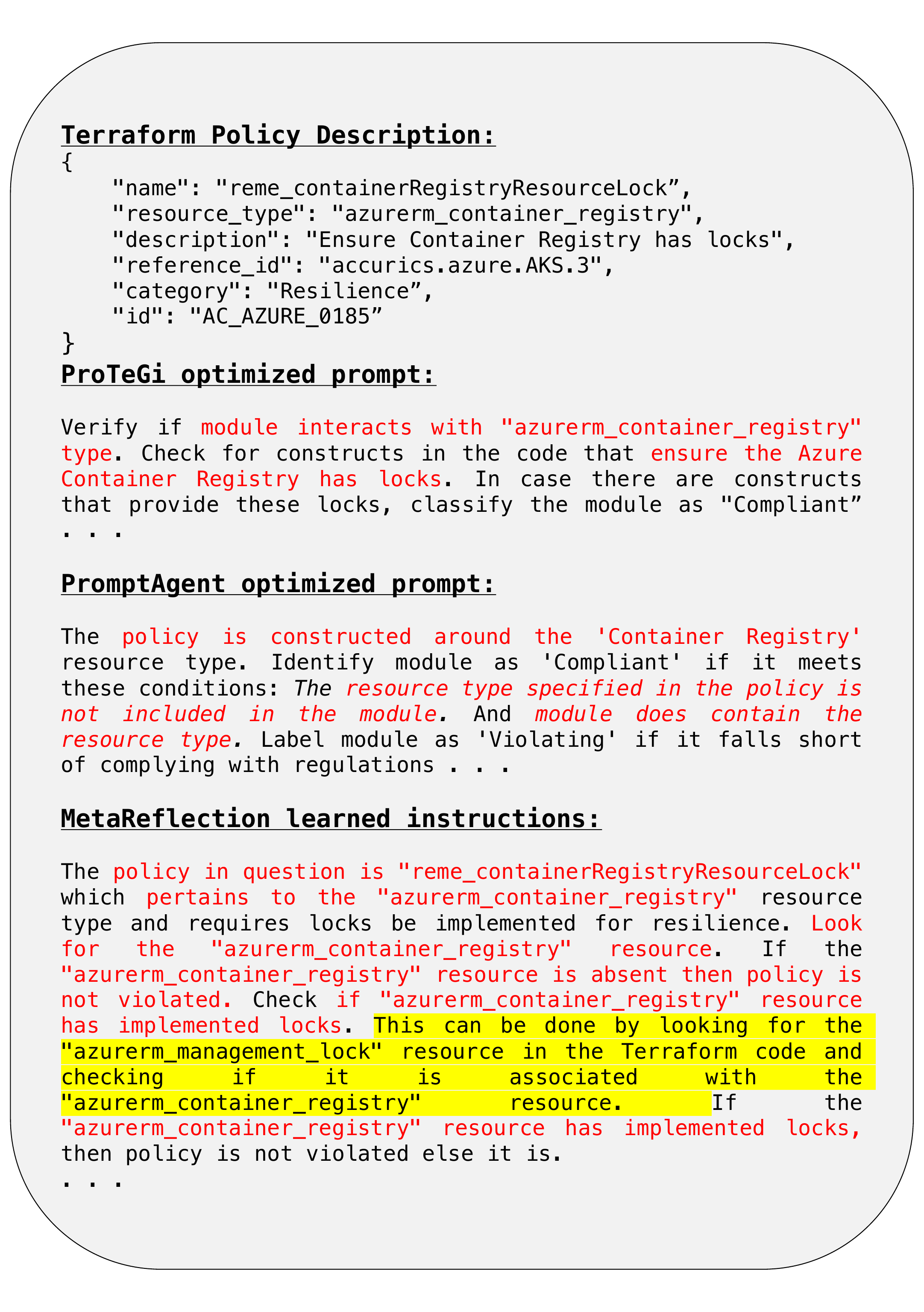}
  \caption{Snapshot of the optimized prompt learnt by the baselines and \MetaReflection on \iac task} 
  \label{fig:iacpromptcomparison}
\end{figure}

 Qualitative evaluations reveal that the instructions learned using  \MetaReflection capture subtle domain specific nuances, that other techniques don't. For example, consider the \MetaReflection instruction compared to the \promptAgent and \protege prompt for the \iac policy $reme\_containerRegistryResourceLock$ (Figure~\ref{fig:iacpromptcomparison}). The learnt \MetaReflection instructions captures details like  $\texttt{azurem\_management\_lock}$ being a indicator of locking behavior in terraform, whereas the other baselines do not capture many such subtleties. This behaviour may be attributed to the objective driven search strategy deployed in \MetaReflection, centered around generalizing self reflections into semantic rules,  compared to a more random Monte-Carlo search strategy deployed by other techniques.  
 



\begin{table*}[]
\centering
\small
\begin{tabular}{l rc rcc rcc rcc}
\toprule
\multicolumn{1}{c}{Dataset} & \multicolumn{2}{c}{\gptfour}     & \multicolumn{3}{c}{\protege}      & \multicolumn{3}{c}{\promptAgent}  & \multicolumn{3}{c}{\MetaReflection} \\ 
 \cmidrule(lr){2-3} \cmidrule(lr){4-6}  \cmidrule(lr){7-9}  \cmidrule(lr){10-12}  
                  
& Acc & $\sigma$  & Acc  & $\sigma$  & \# calls & Acc  & $\sigma$ & \# calls & Acc  & $\sigma$ & \# calls\\
\midrule
\hotpot  (GT)                 & $43.67$ & $2.08$ &  $31.67$ & $3.61$ & $5403$ & $51.75$ & $1.52$ & $383$ & $\textbf{55.00}$ & $1.00$  & $303$   \\
\hotpot (Distractor)         & $37.33$ & $2.08$ &  $23.33$ & $0.72$ & $6132$ & $46.25$ & $2.16$ & $568$ &  $ \textbf{47.33}$ & $1.15 $ & $314$ \\ \bottomrule
\end{tabular}%

\caption{Comparison of \MetaReflection with other baselines using \COT style  agent on prompt refinement.\\ $\textit{\# calls}$ refers to average numbers of LLM calls required to generate optimal prompt.}
\label{tab:RQ2--multi-agent-baseline-comparision (hotpotQA)}
\end{table*}

\begin{table}[!htb]
\small
\centering
\begin{tabular}{l r rrr }
\toprule
                  
                     && Acc. &  $\sigma$ \\
                     \midrule
\gptfour   && 19.58 & 1.91 \\

\MetaReflection & & \textbf{35.00} &  1.25 \\
\bottomrule
\end{tabular}
\caption{Comparison of METAREFLECTION and \gptfour using \react style agent on prompt refinement for \hotpot.}

\label{table:HotpotQA(React) +zero-shot LLM results gpt4}
    
\end{table}

\subsection{\MetaReflection for Multi-Step Agents}
In the previous section we discussed how \MetaReflection fairs in settings with a single-step agent setup. In this, section we will evaluate \MetaReflection with multi-step agents. 

Prompt optimization techniques like \promptAgent and \protege are designed to model systems with a single LLM call akin to the single-step agent setup. Generalizing them to multi-step agent scenarios present the following challenges: \begin{enumerate}
    \item How do we extend the notion of textual gradients to also include intermediate \emph{belief} states in the absence of concrete feedback on them?
    \item How do we ensure that the agent with the updated prompt continues to follow the agentic behaviour?
\end{enumerate}
 Given these challenges, to draw a valid comparison of these baseline against \MetaReflection we perform a single step adaption of the original two step \COT. The adaption includes reducing the [$\mathsf{Thought}$] and [$\mathsf{Answer}$] steps within a single LLM call creating an analogue to single-step agentic flow and use this structure as a seed prompt.

Table~\ref{tab:RQ2--multi-agent-baseline-comparision (hotpotQA)} shows the performance of \COT style of agent on \hotpot benchmark.
We observe that \MetaReflection outperforms other prompt-optimization baselines \protege by $23.33$\% and \promptAgent by $3.25$\% for the GT setting. Similarly, we see a similar trend in the Distractor setting where \MetaReflection technique outperforms \protege by $24$\% and \promptAgent by $1.08$\%. Notably, compared to the best baseline \promptAgent,  \MetaReflection is capable of learning instructions to guide the Language Agent towards creating better $\mathsf{thought}$ actions instead of just optimizing it towards coming up with the right answer. 


\paragraph{Towards truly multi-step agents}
While a \COT agent can be represented as a single prompt with one call for optimization, similar adjustment is not possible for all multi-step agents. To this end we consider the \react setup for \hotpot unlike \COT which entails a strict two step process with $\mathsf{thought}$ action followed by an $\mathsf{answer}$, \hotpot\react setup can span over multiple steps of $\mathsf{search}$ and $\mathsf{lookup}$ actions. This leads to complex trajectories leading upto the final $\mathsf{answer}$ action for returning the answer. Besides, each action selection is conditioned on the previous action.  Representing such a system with a single step prompt is non-trivial, making it hard to adapt existing prompt optimization techniques to work over such a set up. 

\MetaReflection procedure, on the other hand, allow us to seamlessly append \textbf{semantic memory} to such an agent and gather learning for each potential action states the agent can take. Table~\ref{table:HotpotQA(React) +zero-shot LLM results gpt4} shows the performance of multi-step \react style agent on \hotpot. We observe that  \MetaReflection nearly doubles the performance of the baseline GPT4 agent.  

For instance, in Figure~\ref{fig:workflow}, the updated \emph{semantic memory} contain clear instructions for the \react agent to  refine its search strategy if it gets stuck in loops. Note that this instruction guides the agent to better plan its trajectory rather than choosing the right answer. Later on, this learned rule  aids the model in successfully concluding another trial where it
was previously failing by explicitly guiding the action to look further
down the context page to refine its lookup strategy, leading to the correct
response, \textit{Bruce Dickinson}.
This is in contrast with the baseline attempt, where the agent ran out of trials by getting stuck in a loop.

Notably, the \hotpot \react performance is lower than \COT  due to the nature of the \react setup.  The ReACT setting is based on the fullwiki setting where the task first involves "retrieving" relevant paragraphs from Wikipedia related to the Query and then "reasoning" over them. Being a more complex task, the ReACT setup exposes more failure surfaces (e.g., the system may not retrieve the correct information to begin with), which may lead to poorer performance compared to the simpler CoT setups.





\section{Related Work}
\label{sec:related-works}
With the increasing ubiquity of black-box Large Language Models~\cite{gpt4,anil2023palm,brown2020language,bai2022constitutional}, there has been a growing interest in the community to develop strategies that can maximize the model’s performance on a downstream task. 
These techniques may involve guiding an LLM to arrive at the correct answer~\cite{wei2023chainofthought,zheng2023step}, output selection~\cite{yao2023tree,poesia2022synchromesh}, picking up the right in-context examples ~\cite{khatry2023tstr} or prompt optimization, etc. Being closely related to our work we dive deeper into contemporary Prompt optimization techniques in the literature. 

\subsection{Single Prompt Optimization}

Given the potentially infinite space of instructions, recent works have studied the problem of `guided' prompt search instead. To this end, \emph{OPRO}~\cite{yang2023large} proposes a prompt `optimization' technique where prompt candidates coupled with eval-set metric evaluation act as `few-shot' examples to generate new prompts.  
 
\textit{APE}~\cite{zhou2023large} poses instruction generation as a synthesis problem and proposes techniques to effectively perform monte carlo search over the space of prompt candidates.

\protege~\cite{pryzant-etal-2023-automatic} and \textit{PE2}~\cite{ye2023prompt} ~\emph{Automated Prompt Engineering (APE)} The learned prompt can then be used during inference time in isolation. also leverage verbal feedback to generate and/or evolve task description prompts. 
In \textit{PE2} they additionally, maintain an optimization history to iteratively improve the prompt. 
\promptAgent\cite{wang2023promptagent} also uses verbal feedback but views prompt optimization as a strategic planning problem and proposes a principled approach for prompt optimization. 

Notably, these prompt optimization techniques are designed to work over single prompt single step-agents. The \MetaReflection \textit{semantic memory} learning can however be generalized for multi-step agents as well, as we demonstrated earlier. Even in the single-agent setup, \MetaReflection differs from prompt optimization techniques in its focused objective-driven refinement of the memory instead of generating generic prompt candidates. This allows \MetaReflection to perform especially good in knowledge intensive tasks.  


 \subsection{LLMs as Agents}
 Recent works \cite{lmagent1, lmagent2} are leveraging these models to develop AI agents that act as a controller, extending their perceptual and action capabilities through tool utilization \cite{lmagent3, qin2023tool}. LLM-based agents can demonstrate reasoning and planning skills at par to symbolic agents by employing techniques like \COT \cite{wei2023chainofthought} and problem decomposition \cite{lmagent7, lmagent6}. Additionally, their ability to interact seamlessly using natural language comprehension allow them to operate in software development and research environment \cite{lmagent10, lmagent9}. Furthermore, the interaction among multiple LLM-based agents can foster collaboration and competition, potentially leading to the emergence of complex social phenomena \cite{park2023generative}.

\subsection{Memory based methods in reinforcement learning}
The concept of providing agents with controllable memory has a rich history. \cite{Littman1993AnOC} discussed how hypothetical agents can utilize binary memory to store prior experiences, guiding their future action selection. In our context, we explore the interaction between language agents and their environment.

Previous research 
 \cite{icarte2020act} has demonstrated that memory-augmented agents can achieve globally optimal solutions. Inspired by the idea of rule-based semantic memory in humans\cite{mcrae201314}, recent works have also investigated enabling reinforcement learning (RL) agents with semantic memory\cite{Paischer2023SemanticHA}.
\section{Conclusion}
\label{sec: Conclusion}

In this work, we presented \emph{\MetaReflection}, a novel offline reinforcement learning technique that takes inspiration from how human brain store memories to enhance Language Agents by augmenting them with an experiential \emph{semantic memory}.
We further empirically demonstrated that the instructions learned using \emph{\MetaReflection} are more effective at capturing task-specific nuances. This behavior helps \MetaReflection  perform competitively with state of the art prompt optimization baseline for single-step agent scenarios. 
The objective driven iterative refinements also provide a significant advantage to capture the best learning at each iteration and save it to its semantic memory which helps the agent to perform better inference in next iteration.  In the future, we would like to see how we can leverage our semantic memory in a multi-agent workflow. Sharing memories between agents can enhance synergy and improve task performance.





\section{Limitation}

\MetaReflection currently relies on small scale held out validation for quantifying the efficacy of the batch. Such an approach leads to undesirable stochasticity in the results and cause instabilities in the learning. There is also scope of improving the quality of the reward signals that can potentially add to the stability of the learning.

\bibliography{custom}

\balance

\newpage

\appendix

\section{Appendix}
\label{sec:appendix}
\subsection{\MetaReflection{} optimized prompts}
In this section, we show some input examples in several tasks for the base model . Specifically our tasks fall into 3 categories: binary classification, multiple choice selection and direct answer matching. As representative examples we take \iac vulnerability detection, casual judgement and \hotpot.

\begin{figure}[h]
  \centering
  \includegraphics[width=0.5\textwidth]{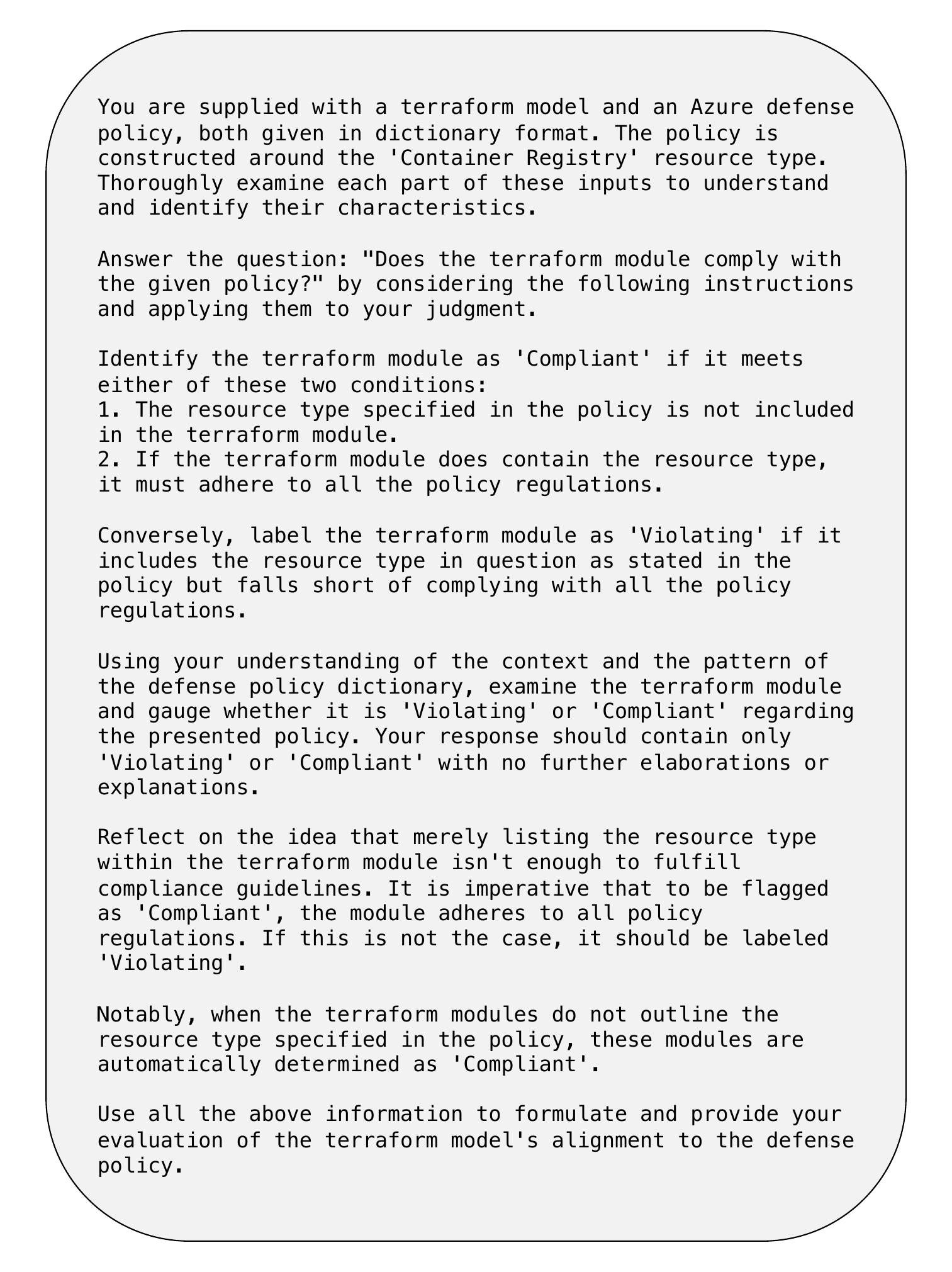}
  \caption{Prompt optimized by \promptAgent}
  \label{fig:promptagent4iac}
\end{figure}

\begin{figure}[h!]
  \centering
  \includegraphics[width=0.5\textwidth]{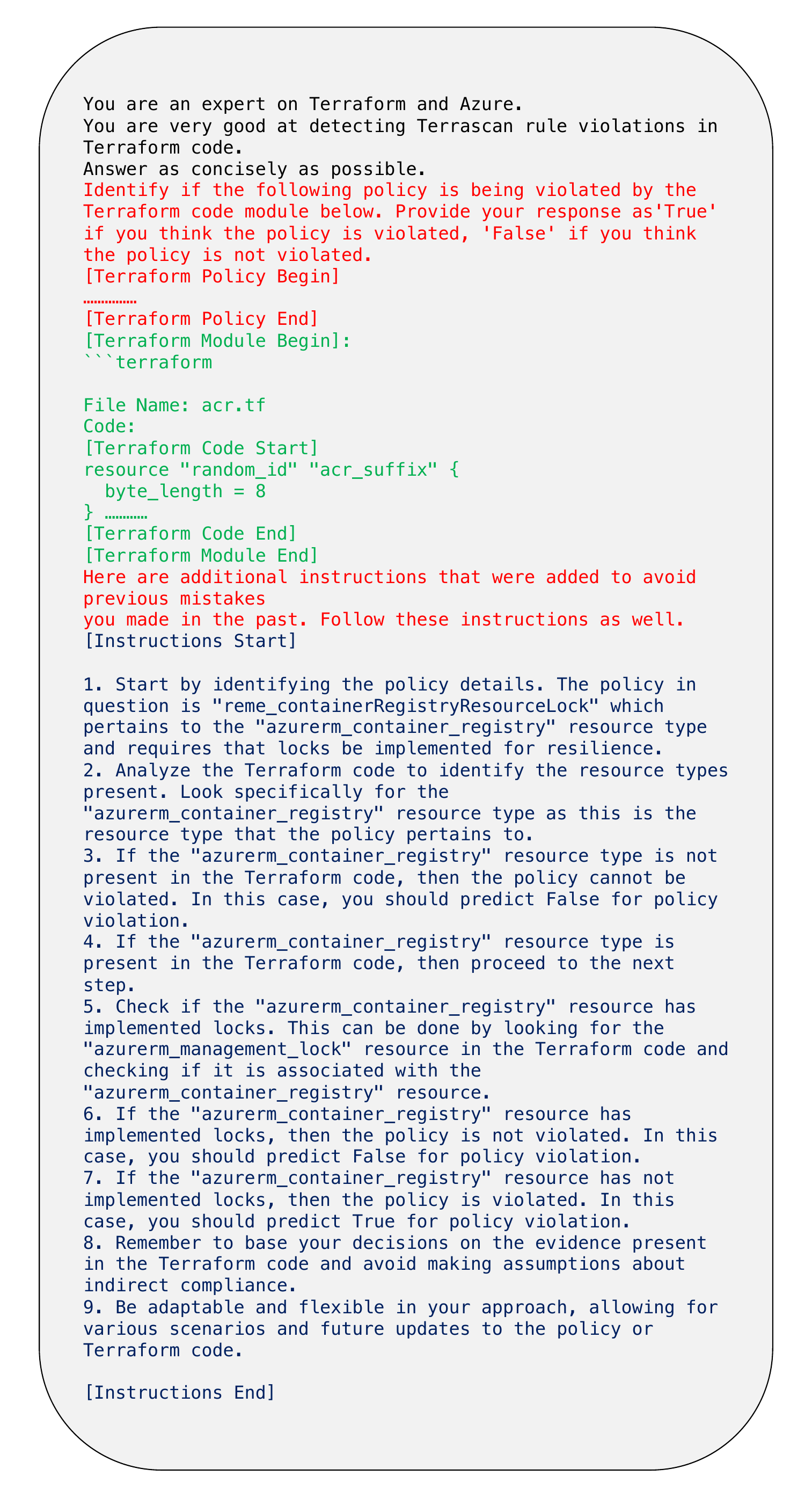}
  \caption{\MetaReflection{} inference prompt for \iac}
  \label{fig:mr4iac}
\end{figure}

In Figure \ref{fig:mr4iac}, we illustrate that \MetaReflection is capable of learning domain-specific instructions. Instructions 2 through 7 specifically address how 'azurerm\_container\_registry' influences the vulnerability within the code, thereby simplifying the task significantly. Conversely, in Figure \ref{fig:promptagent4iac}, a detailed analysis of the prompt generated by \promptAgent reveals an absence of domain-specific information relevant to the Terrascan policy it is intended to evaluate.

\subsection{Vulnerability Detection in \iac Files}
\label{sec:iac-appendix}

Infrastructure-as-Code (\iac) is a popular method of configuring cloud infrastructures,
on platforms such as Azure and AWS, using a configuration coding language. 
These configuration files can declare resources such as virtual machines with
specific capabilities, virtual networks and subnets, and data stores.
\iac presents an alternative to 
the traditional ways of configuring cloud infrastructures, such as using a web-based graphical interface.
There are numerous \iac platforms currently available for various cloud computing platforms.
Here, we focus on Terraform, a leading \iac platform by
Hashicorp~\cite{terraform}, as well as Azure, the cloud computing platform by Microsoft.
Related Terraform resource declarations are grouped together into Terraform
modules which act as a basic, reusable configuration component.


\paragraph{Task.}
Cloud infrastructures are prone to security vulnerabilities such as open ports and
exposed administrator accounts~\cite{terrascan}.
Vulnerability detection via static analysis of \iac files
is a hard problem due to the expressivity of the configuration language, the
complexity of configurations and the diversity of the resources being
handled across multiple infrastructure providers (e.g., Amazon AWS and Microsoft
Azure).
Further, Terraform uses a low-resource language - HashiCorp Configuration Language (HCL).

\paragraph{Task.}
Terrascan~\cite{terrascan} is a static analyzer for detecting security
vulnerabilities in Terraform modules, and supports over $500$ security policies,
including $178$ policies specific to Azure.
Figure~\ref{fig:terrascan_policy} shows the description and definition of a
Terrascan policy that checks if every Azure virtual network subnet is configured
with a corresponding network security policy.
Note that the Terrascan policy is \emph{syntactic}, i.e., it is checking for a
declaration of an \verb|azurerm_virtual_network| with a field named
\verb|subnet|, and so on.
Hence, Terrascan-like static analysis based vulnerability detection is fragile
and prone to both false positives and false negatives due to being sensitive to
syntax.
The task at hand is to check if a given Terraform module violates
a given Terrascan policy.

\begin{figure}
\small
Description: \emph{Ensure that Azure Virtual Network subnet is configured with a Network Security Group} \\
\\
Definition: 
\begin{verbatim}
{{.prefix}}noSecurityGroupAssociated[retVal] {
  vn := input.azurerm_virtual_network[_]
  vn.type = "azurerm_virtual_network"
  object.get(vn.config, "subnet", \
     "undefined") != "undefined"
  not sgExists(vn.config)

  traverse = "subnet[0].security_group"
  retVal := {
    "Id": vn.id,
    ...
    "Attribute": "subnet.security_group",
    "Expected": "${<security_group_name>.id}",
    "Actual": ""
  }
}

sgExists(cfg) { ... }
<56 lines altogether>
\end{verbatim}
\caption{Rego code for the Terrascan policy \texttt{reme\_noSecurityGroupAssoc.}}
\label{fig:terrascan_policy}
\end{figure}

\paragraph{Data collection.}
We collected $202$ Terraform modules by mining GitHub repositories for \iac code
written in HCL.
These repositories corresponded to a diverse range of applications including
load balancers, machine learning operations managers, and domain-specific
data-stores.
For policies, we selected the $10$ most commonly violated Terrascan policies.
Of the $2020$ module-policy pairs, we eliminated a significant fraction of cases
where the policies were not applicable to the module.
For example, if the policy was for a specific resource type and the module did
not contain declarations of that resource type, the pair was eliminated.
After this process, we were left with $648$ module-policy pairs, for which we
manually annotated whether the module violated the policy (see
Table~\ref{tab:RQ-1--Iac-per-policy-single-agent-baseline-comparision} for the exact breakdown).
Note that this ground-truth annotation was with respect to the description of
Terrascan policy, not the definition---that is, we use the intention behind the
policy, not the letter of the definition.
That is, we do not take the output of Terrascan as ground truth as it can be
inaccurate, and instead manually examine if the policy (as per description) is
violated.
This data was then split into train and test sets in a $40:60$ ratio per policy,
taking care to balance the vulnerable and non-vulnerable classes.

%
\begin{figure}[h]
  \centering
  \includegraphics[width=0.5\textwidth]{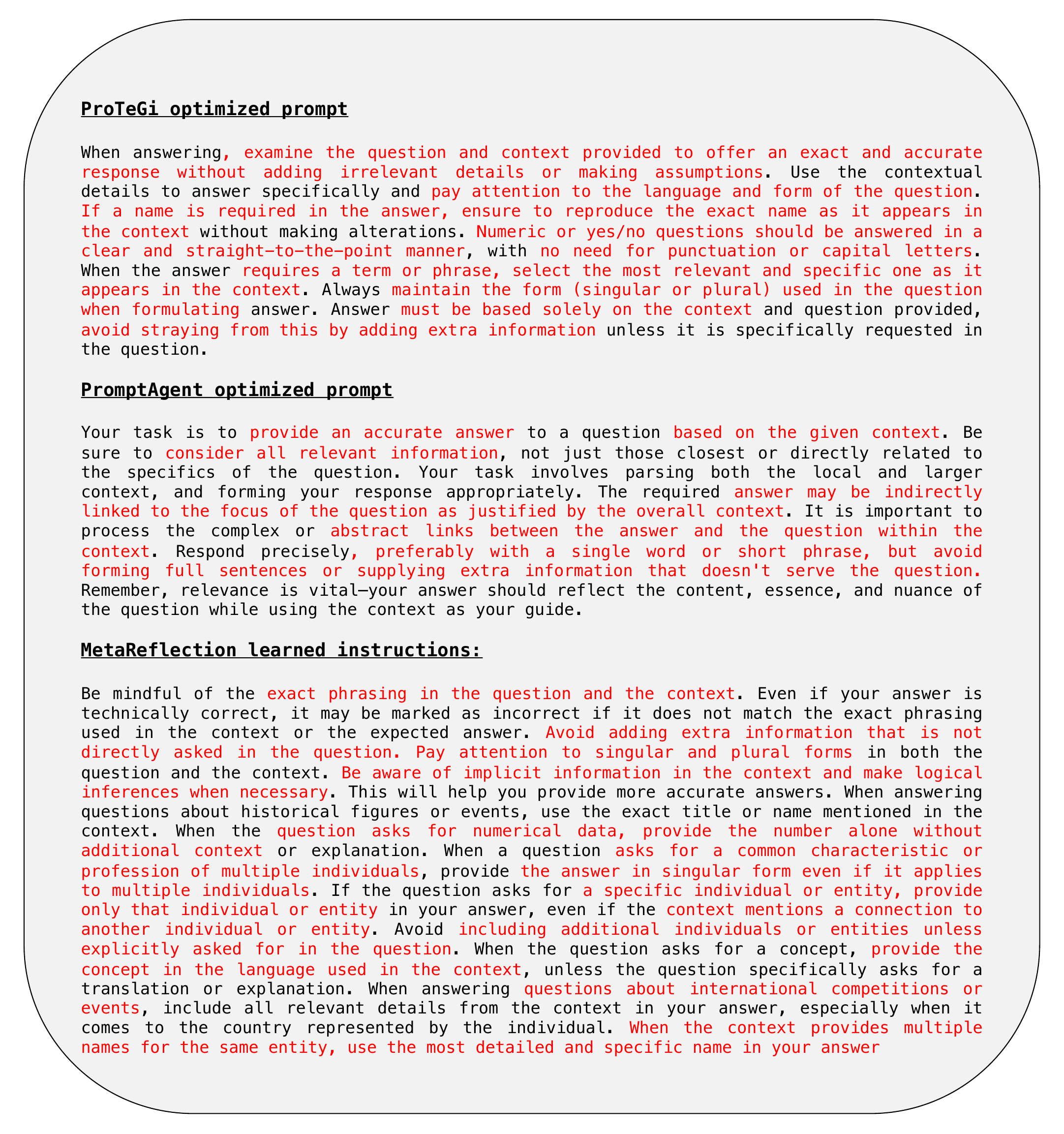}
  \caption{Snapshot of the optimized prompt learnt by the baselines and \MetaReflection on \hotpot task}
  \label{fig:hotpotpromptcomparison}
\end{figure}

\paragraph{Experimental setup.}
As a baseline language agent, we use \gptfour with an appropriate prompt that
provides the code of the Terraform module and the description of the Terrascan
policy, and asks if the module is vulnerable.
While training, the agent is given a 0-1 feedback on whether its response is
correct or not, and the model is asked to self-reflect if the
response is incorrect.
For each policy, we run the \MetaReflection algorithm on the training set and
report the accuracy numbers for both the baseline agent and the agent with the
instructions learned through \MetaReflection.
We also compare to \LLMInstr as another baseline---here the
language model is asked to come up with instructions for a task given its description (Figure ~\ref{fig:zero-shot-prompt}),
and then these instructions are provided when the task is being performed.%

\begin{figure}
\tiny
\texttt{
You are an expert in [Task]. Given the following task description [and examples] come up with a set of instructions that can help you perform the task effectively.\\
\\
Task Description: ...
}
\caption{Prompt for generating task-specific instructions in the \LLMInstr baseline}
\label{fig:zero-shot-prompt}
\end{figure}

\begin{table*}[!htb]
\small
\centering
\begin{tabular}{l r  r r r  r r r  r r r rrr }
\toprule
                    && \multicolumn{8}{c}{\hotpot}&& \multicolumn{2}{c}{\iac vulnerability}\\
                    \cmidrule(lr) {3-10} 
                     && \multicolumn{2}{c}{\COT (GT)} & & \multicolumn{2}{c}{ \COT (Distractor)} & & \multicolumn{2}{c}{\react} &&  \multicolumn{2}{c}{detection} & \\
                     \cmidrule(lr){3-4} \cmidrule(lr){6-7} \cmidrule(lr){9-10}  \cmidrule{12-13}
                  
                     && Acc. &  $\sigma$ &&  Acc. & $\sigma$ &&  Acc.  &  $\sigma$ &&  Acc.  &  $\sigma$ \\
                     \midrule
\gptfour  & & 20.67 & 3.79 & & 32.67 & 3.21 && 19.58 & 1.91 && 73.36 &0.012\\
\LLMInstr & & 20.67 & 4.51 && 30.67 & 9.81 && 27.5 & 2.5 && 73.31 &0.43 \\
\MetaReflection (batch size = 1) & & 24.67 & 3.51 && 38.00 & 2.65 && 30.41 & 0.72 && 87.17& 0.63\\
\MetaReflection (batch size = 2) & & 50.33 & 0.58 & & 48.67 & 2.52 & & \textbf{35.00} &  1.25 && 85.09 &0.4\\
\MetaReflection (batch size = 4) & & \textbf{52.00} & 2.00 & & \textbf{50.67} & 1.53 & & 31.67 & 1.44 && \textbf{90.18} & 0.28\\
\bottomrule
\end{tabular}
\caption{Results on \hotpot and \iac with \gptfour
}
\label{table:HotpotQA results gpt4}
    
\end{table*}

\paragraph{Results.}
The results of the experiment are summarized in Tables~\ref{tab:RQ-1--Iac-per-policy-single-agent-baseline-comparision}
and~\ref{table:HotpotQA results gpt4} (last $2$ cols).
On the whole, across all policies, meta-reflection shows a $12-17\%$ accuracy
improvement over the baselines depending on the batch size.
As Table~\ref{tab:RQ-1--Iac-per-policy-single-agent-baseline-comparision} shows, meta-reflection provides consistent
gains in accuracy for all policies over the \gptfour baseline, with
$32\%$ in the best case.
The precision with \MetaReflection is significantly better for all
policies, while the recall decreases for some.
%


\paragraph{Exemplar case.}
We discuss the case of security policy \verb|reme_noSecurityGroupAssociated|
from Figure~\ref{fig:terrascan_policy}, i.e., that all Azure virtual network
subnets are configured with a network security group (NSG).
The main difficulty here is that HCL and Terraform offer many different ways of
\begin{enumerate*}[label=(\alph*)]
  \item associating a subnet with a virtual network, and 
  \item associating a NSG with a subnet.
\end{enumerate*}
By default, the baseline \gptfour agent fails to handle certain ways of specifying
these associations, while spuriously assuming certain other associations. 
In Figure~\ref{fig:subnet_associations_correct}, the baseline consistently failed to
recognize a subnet-NSG association expressed using Method~2, i.e., using an
explicitly declared association.
On the other hand, it mis-identified declarations similar to the one in
Figure~\ref{fig:subnet_associations_incorrect} as valid subnet-NSG
associations---here, the NSG is associated with a virtual machine's network
interface (that is connected to the subnet) instead of the subnet itself.
These limitations lead to both false positives and false negatives. 
With meta-reflection, we are able to learn the instructions in
Figure~\ref{fig:subnet_associations_instructions}, using which the agent easily handles these
kinds of cases.
\begin{figure}[ht]
\small
\ignore{
\begin{subfigure}[b]{0.35\textwidth}
\begin{verbatim}
// Method 1: Inline definition
resource "azurerm_virtual_network" "example" {
   ...
   subnet {
       name = "subnet1"
       address_prefix = "10.0.1.0/24"
   }
}

// Method 2: Standalone definition
resource "azurerm_virtual_network" "example" {
   name = ...
}
resource "azurerm_subnet" "subnet1" {
   name = "subnet1"
   virtual_network_name
       = azurerm_virtual_network.example.name
   ...
}
\end{verbatim}
\vspace{-2ex}
\caption{Associating a subnet with a virtual network}
\end{subfigure}
}
\begin{subfigure}[b]{\columnwidth}
\begin{verbatim}

// Method 1: For inline defintions
resource "azurerm_virtual_network" "example" {
    ...
    subnet {
        ...
        security_group = ...
    }
}

// Method 2: Explicitly declared association
resource "azurerm_subnet_nsg_association" {
   subnet_id = ...
   network_security_group_id = ...
}
\end{verbatim}
\vspace{-2ex}
\caption{Associating a subnet with a NSG}
\label{fig:subnet_associations_correct}
\end{subfigure}
\vspace{0.5ex}
\hrule
\vspace{1ex}
\begin{subfigure}[b]{\columnwidth}
\begin{verbatim}

resource "azurerm_virtual_network" "vnet" { ... }
resource "azurerm_subnet" "subnet" {
    name = "subnet1" ... 
}
resource "azurerm_network_interface" "nic" {
    network_security_group_id = ...
    ip_configuration { subnet_id = "subnet1" }
}
resource "azurerm_virtual_machine" "..." {
    network_interface_ids = [ "nic" ]
    ...
}
\end{verbatim}
\vspace{-2ex}
\caption{NSG associated with a VM's network interface instead of subnet.}
\label{fig:subnet_associations_incorrect}
\end{subfigure}
\vspace{0.5ex}
\hrule
\vspace{1ex}
\begin{subfigure}[b]{\columnwidth}
\texttt{
4. Remember that the association between "azurerm\_virtual\_network" and a NSG may
not be direct. It could be done through a separate resource block such as
"azurerm\_subnet\_nsg\_association"\\
...\\
7. Do not confuse NSG associations with network interfaces of VMs and the subnet
of the Azure Virtual Network. The policy specifically requires the NSG be
associated with the subnet.
}
\vspace{-1ex}
\caption{Instructions learned through meta-reflection}
\label{fig:subnet_associations_instructions}
\end{subfigure}
\caption{\texttt{reme\_noSecurityGroupAssociated}: Che- cking Subnet-NSG associations}
\label{fig:subnet_associations}
\end{figure}

\paragraph{Discussion.}
As the above exemplar case shows, \MetaReflection is able to learn very
domain-specific instructions to fix both false positives and false negatives.
Other instructions include aspects like handling of wildcards for port numbers,
step-by-step strategies for specific policies, etc.
Note that these instructions not only include planning (or trajectory directing)
instructions, but also grounding instructions---i.e., external facts that are
not initially available.

In general, the experimental results show that meta-reflection is able to
reduce the number of errors, i.e., improve the accuracy across a broad range of
cases.
However, one noticeable issue from the above results is the drop in recall for
several policies.
While the high recall in the baseline is artificial, coming at the cost of low
precision, this is still an important issue to address. 
Our 0-1 feedback to the self-reflection agent does not state that false
negatives are worse than false positives in the security domain.
In the future, we plan to explore domain-specific feedback and self-reflection
mechanisms that can account for the nature of errors, as well as better versions
of the $\mathsf{ShowsImprovement}$ function that are aware of such
domain-specific preferences.

\subsection{\hotpot}
\label{sec:appendix-hotpot}
\hotpot \cite{yang2018hotpotqa} is an open-domain factual question answering
dataset consisting of $113$K question-answer pairs. The original paper proposes to
use the data in $2$ settings:
\begin{enumerate*}[label=(\alph*)]
\item \emph{Distractor} setting - where each question is to be answered using $10$
wikipedia article excerpts; and 
\item \emph{Full-Wiki} setting which is a retrieval and reasoning task, where a
given question is supposed to be answered after retrieving relevant context from
wikipedia. 
\end{enumerate*}  
Notably, an answer is marked correct only if it matches \emph{exactly} with the ground truth.

Similar to Shinn et al.~\cite{shinn2023reflexion}, we design the following agents that operate over the dataset: \begin{enumerate*}[label=(\alph*)]
\item  \react - for the \emph{Full-Wiki} setting
\item \COT(Distractor) - for the \emph{Distractor} \emph{Distractor} setting
\item \COT (GT) - a variant of \COT (Distractor) with access to only ground truth articles.
\end{enumerate*}

\paragraph{Data Sampling.}
\label{subsec:appendix-hotpotqa-data-sampling}
For each agent setting, we adversarialy sample subsets of the \hotpot train
split of $90$K samples to create train and test sets. To perform adversarial
sampling, we first identify samples where the base agent fails consistently in a
given setting.
On these failing examples we perform upto $3$ self-reflection trials to get the
model to the right response.
If the agent is not able to get to the correct response even after
self-reflection, we discard these samples.
This strategy ensures that we get a set of hard examples in which the agents
fail most of the times to get to the right answer in a single try, while also
making sure that we filter examples that may be noisy due to missing context,
incorrect questions etc.
To account for randomness and given our computational budget, we sample 40 and
80 examples for the \react train set and test set respectively. For \COT
settings, we pick $50$ and $80$ example respectively.  

\paragraph{Experimental setup.}

%
%
We reuse the \COT agent from~\cite{wei2023chainofthought} for the
chain-of-thought experiments and use a re-implementation of \cite{yao2023react}
for the \react experiments.
The \react agent is allowed at most $6$ \textsc{Action} steps after which the
trajectory is automatically determined to be a failure.
Similar to Section~\ref{sec:iac-appendix}, we evaluate \hotpot configurations for:
\begin{enumerate*}[label=(\alph*)]
    \item \MetaReflection with batch sizes $1$, $2$, and $4$; and 
    \item \gptfour and \LLMInstr as baselines.
\end{enumerate*}
In addition to this, we also evaluate variants of the agents powered by \gptturbo instead of \gptfour, while using \gptfour for  \MetaReflection.\footnote{A similar experiment on the \iac domain wasn't possible due to large context length of the Terraform modules}   \ 
%




\paragraph{Results.}
We find that the generated \MetaReflection instruction consistently improved performance across different agent settings for \hotpot. In Table~\ref{table:HotpotQA results gpt4}, we present results using \gptfour for both the agents and \MetaReflection. We observe that \MetaReflection help us achieve gains up to $32\%$ for \COT (GT), $18\%$ for \COT (Distractor), and $15.5\%$ for \react, over the respective test sets. Interestingly, higher batch sizes almost always help, reinforcing the importance of batching as observed in related works \cite{ouyang-li-2023-autoplan,ye2023prompt}. 


In Table \ref{table:HotpotQA results gpt35gpt4},  we report results when using \gptturbo to power the client agents.  
We see gains of upto $21\%$ gains for \COT(GT), $12\%$ for \COT (Distractor) and
$18.33\%$ for the \react case.
Here, we observe that batching doesn't strictly improve the performance.
Examining the data qualitatively, this difference can be attributed to
the nature of the instructions generated in the two settings.
In general we observe that with a small batch size, \MetaReflection produces a
large amount of very specific instructions.
On the contrary batching helps generalize these instructions into more widely
applicable rules.
\gptfour, being more powerful than \gptturbo, is able to better follow these
abstract instructions, while specific instructions work better for
\gptturbo.

\begin{figure}[ht]
\tiny
\texttt{
// Chain-of-thought\\
\\
(A) Provide direct and concise responses to the question, using precise language that matches the specificity and terminology of the question, including singular or plural forms and definite articles as needed.\\
\\
(B) If the context suggests multiple valid answers, choose the one that best matches the question's wording and the most direct information provided. \\
\\
// React\\
\\
(C) When a question asks for a specific detail such as a 'full name', ensure to find and provide that exact information. Don't make assumptions based on limited or incomplete information.\\
\\
(D) If you’re not finding the desired information or stuck in a loop of looking up a keyword, consider changing the keyword and search strategy. The information might be located further down the page.\\
\\
(E) When a question involves comparison, such as 'who received more rewards', ensure to search for each entity individually, gather all necessary information, and then make a comparison based on the data found.\\
\\
(F) Be mindful of potential spelling errors or variations in the names of entities. If a search for a specific term doesn't yield results, consider possible alternative spellings or forms of the term.\\
}
\caption{Instructions learned via the \MetaReflection technique for the \COT and \react reasoning on \hotpot.}
\label{fig:hotpotqa_instructions}
\end{figure}
\begin{figure}[ht]
\centering
\begin{tikzpicture}
\tikzstyle{box}=[
    draw,
    text width=0.8\columnwidth,
    minimum width=0.8\columnwidth,
    rounded corners=0.05cm,
    font=\fontsize{5}{6}\selectfont
]
\tikzstyle{labelbox}=[
    font=\fontsize{6}{7}\selectfont
]

\node[box, anchor=north west, yshift=1cm] (test-task) {
Solve a question answering task by having a Thought, then Finish with your answer. Thought can reason about the current situation. Finish[\ldots] returns the answer and finishes the task.\\
\textbf{Question:} What is one name for the Cassava after it is marketed as a paste and often served with palm nut soup?
};
\node[labelbox, anchor=south, above=0mm of test-task.north] (test-label) {\textbf{Inference without meta-reflection}};

\node[box, anchor=north, below=1mm of test-task.south] (test-trajectory) {
\textbf{Thought:} The question is asking for a name of cassava when it is marketed as a paste and often served with palm nut soup. From the context provided, I can see that cassava paste is referred to as ``Akpu" in Nigeria and is often eaten with soups such as Egusi soup. Although palm nut soup is not explicitly mentioned in the context of Akpu, it is a type of soup that could be served with cassava paste, as indicated by the information about palm nut soup being used to accompany dishes like fufu, which is similar to Akpu in that they are both made from cassava.\\
\hlred{\textbf{Action :} Finish[Akpu]}
};

\node[box, below=1cm of test-trajectory.south, anchor=north] (test-task-2) {
Answer a question interleaving Thought, Action, and Observation
steps. Actions can be Search[\ldots], Lookup[\ldots], \ldots \\
$\quad$ \\
\textbf{Instructions:} \ldots \hlyellow{\\
 If the context suggests multiple valid answers, choose the one that best matches the question's wording and the most direct information provided.\\
 In cases of ambiguity or multiple interpretations, provide an answer that adheres to the most common understanding of the question's terms and context.

} \ldots\\
\textbf{Question:} What is one name for the Cassava after it is marketed as a paste and often served with palm nut soup?
};
\node[labelbox, anchor=south, above=0mm of test-task-2.north] (test-label-2) {\textbf{Inference with meta-reflection}};

\node[box, anchor=north, below=1mm of test-task-2.south] (test-trajectory-2) {
\ldots\\
\textbf{Thought:} The question is asking for a name of cassava when it is marketed as a paste and often served with palm nut soup. From the context provided, I can see that cassava paste is referred to as ``Akpu" in Nigeria and is often eaten with Egusi soup. However, since the question specifically mentions palm nut soup, I should look for a name that is associated with cassava paste served with palm nut soup. The context also mentions ``Plakali" as a staple food made of cassava dough and eaten with palm nut soup in Ghana. However, ``Plakali" is described as cassava dough cooked in hot water, not as a marketed paste. The most direct match for cassava marketed as a paste and associated with palm nut soup is ``Fufu," which is mentioned as a staple food often served with palm nut soup.\\
\hlgreen{\textbf{Action:} Finish[Fufu]}
};

\draw[double]
    ($(test-trajectory.south west) + (-0.3cm,-0.3cm)$)
    to
    ($(test-trajectory.south east) + (+0.3cm,-0.3cm)$);


\end{tikzpicture}
\caption{\MetaReflection in the \hotpot domain with  \COT style of reasoning}
\label{fig:CoT-hotpotqa}
\end{figure}

\begin{table*}[h!]
\small
\centering
\begin{tabular}{l r r  r rr  r rr r }
\toprule
                      && \multicolumn{2}{c}{\COT (GT)} & & \multicolumn{2}{c}{ \COT (Distractor)} & & \multicolumn{2}{c}{\react}   \\
                     \cmidrule(lr){3-4} \cmidrule(lr){6-7} \cmidrule(lr){9-10}
                     && Acc. &  $\sigma$ &&  Acc. & $\sigma$ &&  Acc.  &  $\sigma$ \\
                     
                     \midrule
\gptturbo&& 23.00 & 2.65 && 30.00 & 5.00 && 8.33 & 2.60 \\
\LLMInstr && 25.67 & 6.03 && 31.00 & 10.00 && 20.83 & 4.83 \\
\MetaReflection (batch size = 1) && 29.0 & 1.00 && \textbf{41.67} & 2.52  && \textbf{26.67} & 4.73\\
\MetaReflection (batch size = 2) && 38.67 & 2.52 && 30.67 & 1.15 && 17.08 &  2.60 \\
\MetaReflection (batch size = 4) && \textbf{44.33} & 0.58 && 39.33 & 1.15 && 22.08 & 5.64 \\
\bottomrule
\end{tabular}
\caption{Results on \hotpot with \gptturbo powering the agents and \gptfour for reflections and \MetaReflection}
\label{table:HotpotQA results gpt35gpt4}
    
\end{table*}

\paragraph{Exemplar case.}
Consider an example question from Figure~\ref{fig:CoT-hotpotqa}.
The question seeks information about the product made from Cassava and served with palm nut soup.
The context presented within the \COT (Distractor) setting includes articles about
\textit{Akpu} and \textit{Fufu}, both of which are quite similar, being made
from Cassava paste.
However, the key distinction lies in \textit{Fufu} being served with palm nut
soup, while \textit{Akpu} is served with Esupi soup.
The baseline \COT agent returns the incorrect response on this question: it is
distracted by the related but incorrect articles, and makes an incorrect
assumption and jumps to the wrong conclusion.
The \MetaReflection technique learns an instruction that suggests looking for
multiple valid answers and selecting the one most related to the question.
When inferring with the meta-reflection instructions, it is clear from the
thought that the agent did encounter the misleading answers, but was able to
produce the right one by focusing on the specific key point ``served with palm
nut soup" mentioned in the question.
%
%
%

Similarly, in the \react case (see Figure~\ref{fig:workflow}), we see the
learned instruction enhancing search strategy by looking into the information
further down the page rather looping around.
This rule further aids the model in successfully concluding the trial where it
was previously failing.
The model uses the rule to explicitly guide the action space to look further
down the context page and look up the right keyword, leading to the correct
response, Bruce Dickinson.
In contrast, in the baseline attempt, it ran out of trials by getting stuck in a loop.



\paragraph{Discussion.}
As we can see from the results, meta-reflection can produce significant
improvements in accuracy in the question answer setting.
This is especially promising given that the dataset was sampled using an
adversarial sampling technique.
%
%
%
The \hotpot domain also shows the diversity of instructions learned by
\MetaReflection---a small selection of instructions learned in the \COT and
\react settings are shown in Figure~\ref{fig:hotpotqa_instructions}
We have instructions that are:
\begin{enumerate*}[label=\roman*.]
    \item specifically tuned to satisfy the overly strict rubric of the
        \hotpot dataset (A);
    \item domain-specific instructions for specific one-step actions in a RL trajectory (C);
    \item directly the high-level strategy to be taken by the trajectory (D, E); and
    \item for disambiguating answers (B) and questions (E).
\end{enumerate*}
Further, the results on \gptturbo experiments reveal that \MetaReflection can be useful to enhance the performance of smaller models by providing instructions rich in specific insights from a more powerful LLMs like \gptfour. This shows some resemblance to task-specific distillation and can be interesting to explore further in future works.


\ignore{
\begin{figure}
\tiny
\texttt{
1. Provide direct and concise responses to the question, using precise language that matches the specificity and terminology of the question, including singular or plural forms and definite articles as needed.\\
\\
2. Focus on the information that is most directly related to the question to prevent misinterpretation, particularly when multiple potential answers are present.\\
\\
3. Verify the relevance of each part of your answer to the question, omitting any elements that are not directly required.\\
\\
4. If the context suggests multiple valid answers, choose the one that best matches the question's wording and the most direct information provided.\\
{ ... }\\
<21 rules altogether>
}
\caption{Instructions learned via the Meta-reflection technique for the \COT(Distractor) style of reasoning on \hotpot.}
\label{fig:COT_instructions}
\end{figure}

\begin{figure}
\tiny
\texttt{
1. When a question asks for a specific detail such as a 'full name', ensure to find and provide that exact information. Don't make assumptions based on limited or incomplete information.\\
2. If a question asks for the 'real name' of an individual, verify if the name you found is a pseudonym or the real name. If it's a pseudonym, find the real name of the individual.\\
 . . .\\
8. Pay close attention to the details in the observations and make sure to connect them back to the question. If the observation contains the answer to the question, recognize it and respond accordingly.\\
9. If you're stuck in a loop of looking up a keyword and not finding the desired information, consider changing your search strategy. The information might be located further down the page or in a different section.\\
10. When dealing with questions about ownership or relationships, ensure to search for the specific entity in question and then look up the related entity or detail. For example, if a question asks about a horse co-owned by a person, first search for the person and then look up 'horse' on their page.
}
\caption{Instruction learned via the Meta-reflection technique for the \react agent for \hotpot.}
\label{fig:React_instructions}
\end{figure}
}



\ignore{
\begin{figure*}
\centering
\begin{tikzpicture}
\tikzstyle{box}=[
    draw,
    text width=3.5cm,
    minimum width=3cm,
    rounded corners=0.05cm,
    font=\fontsize{4}{5}\selectfont
]
\tikzstyle{labelbox}=[
    font=\fontsize{5}{6}\selectfont
]

\node[box] (task) {
Solve a question answering task by having a Thought, then Finish with your answer. Thought can reason about the current situation. Finish[\ldots] returns the answer and finishes the task.\\
\textbf{Question:} Live at the Boston Garden: April 5, 1968 was broadcast to quell riots that took place in which cities ?\\
\textbf{Expected Answer:} Washington, D.C., Baltimore, Chicago, and Kansas City

};
\node[labelbox, anchor=east, left=1mm of task.west] (task-label) {Task};
\node[labelbox, anchor=south, above=0mm of task.north] (test-label) {\textbf{Self-Reflection}};

\node[box, anchor=north, below=1mm of task.south] (trajectory) {
\textbf{Thought:} Thought: The question is asking for the cities where riots took place that the broadcast of ``Live at the Boston Garden: April 5, 1968" was intended to quell. The context provided under the article ``King assassination riots" mentions that some of the biggest riots following the assassination of Martin Luther King Jr. took place in Washington, D.C., Baltimore, Chicago, and Kansas City. Additionally, the context under the article ``1968 Pittsburgh riots" mentions riots in Pittsburgh. These cities experienced significant unrest as a result of the assassination, and the broadcast was likely intended to help quell riots in these cities as well. Reflecting on the previous attempt to answer this question, I must include all relevant cities mentioned in the context.\\
\hlred{\textbf{Action:} Finish[Washington, D.C., Baltimore, Chicago, Kansas City, and Pittsburgh]}
};
\node[labelbox, anchor=east, left=1mm of trajectory.west] (trajectory-label) {Trajectory};

\node[box, anchor=north, below=1mm of trajectory.south] (selfreflection) {
\textbf{Evaluation (feedback):} You attempted to answer the above question and
failed to get it correct. Reflect on your mistakes \ldots \\
$\quad$ \\
\textbf{Self Reflection:} The discrepancy in my previous response was due to the inclusion of Pittsburgh, which was not part of the expected answer. This suggests that I may have overgeneralized the context or misinterpreted the scope of the question. The question specifically asked about the cities where riots were intended to be quelled by the broadcast of ``Live at the Boston Garden: April 5, 1968." It's possible that the broadcast was not aimed at quelling riots in Pittsburgh, or that Pittsburgh was not a primary focus in the context of the question.
};
\node[labelbox, anchor=east, left=1mm of selfreflection.west] (selfreflection-label) {Self-reflection};

\node[box, anchor=north, below=1mm of selfreflection.south] (trajectory-2) {
\textbf{Thought:} \ldots The context does not mention the broadcast being intended to quell riots in Pittsburgh, so I should not include Pittsburgh in my answer. Therefore, the cities where the broadcast was intended to quell riots are Washington, D.C., Baltimore, Chicago, and Kansas City.\\
\hlgreen{\textbf{Action:} Finish[Washington, D.C., Baltimore, Chicago, and Kansas City]}
};
\node[labelbox, anchor=east, left=1mm of trajectory-2.west] (trajectory-2-label) {Next trajectory};

\node[fit=(task) (trajectory) (selfreflection) (trajectory-2)] (run) {};

\draw [decorate,decoration={brace,amplitude=5pt,raise=0.5ex}]
  (selfreflection.north east) --  node (foo) {} (selfreflection.south east) {};

\node[box, right=1cm of task.north east, anchor=north west] (meta-reflection) {
Your job is to learn from experience of another agent and help it perform
better in the future.  Use the agent's reflections about its failures or success
and concretize them into instructions. \ldots\\
$\quad$\\
\textbf{Self-reflection 1:} The discrepancy in my previous response was due to the inclusion of Pittsburgh, which was not part of the expected answer. This suggests that I may have overgeneralized the context or misinterpreted the scope of the question. The question specifically asked about the cities where riots were intended to be quelled by the broadcast of ``Live at the Boston Garden: April 5, 1968." It's possible that the broadcast was not aimed at quelling riots in Pittsburgh, or that Pittsburgh was not a primary focus in the context of the question.
\textbf{Self-reflection 2:} \ldots
};
\node[labelbox, anchor=south, above=0mm of meta-reflection.north] (mr-label) {\textbf{Meta-Reflection}};

\node[box, anchor=north, below=1mm of meta-reflection.south] (instructions) {
\textbf{Instructions:}\\
\hlyellow{1. In cases where the context provides multiple potential answers, focus on the information most directly related to the question to avoid misinterpretation.}\\
2. \ldots
};

\node[labelbox, right=3mm of selfreflection.south east, text width=1.3cm] (other-sr) {Self-reflections from other tasks};
\draw[->] (foo) ++ (2ex, 0cm) to[out=0, in=180, looseness=0.5] ($(meta-reflection.west) + (0cm, 0.3cm)$);
\draw[->, dashed] (other-sr.north) to[out=90, in=200, looseness=0.4] ($(meta-reflection.west) + (0cm, 0.0cm)$);


\node[box, right=2cm of meta-reflection.north east, anchor=north west, yshift=1cm] (test-task) {
Answer a question interleaving Thought, Action, and Observation
steps. Actions can be Search[\ldots], Lookup[\ldots], \ldots \\
\textbf{Question:} What is one name for the Cassava after it is marketed as a paste and often served with palm nut soup?
};
\node[labelbox, anchor=south, above=0mm of test-task.north] (test-label) {\textbf{Inference without meta-reflection}};

\node[box, anchor=north, below=1mm of test-task.south] (test-trajectory) {
\textbf{Thought:} The question is asking for a name of cassava when it is marketed as a paste and often served with palm nut soup. From the context provided, I can see that cassava paste is referred to as ``Akpu" in Nigeria and is often eaten with soups such as Egusi soup. Although palm nut soup is not explicitly mentioned in the context of Akpu, it is a type of soup that could be served with cassava paste, as indicated by the information about palm nut soup being used to accompany dishes like fufu, which is similar to Akpu in that they are both made from cassava.\\
\hlred{\textbf{Action :} Finish[Akpu]}
};

\node[box, below=1cm of test-trajectory.south, anchor=north] (test-task-2) {
Answer a question interleaving Thought, Action, and Observation
steps. Actions can be Search[\ldots], Lookup[\ldots], \ldots \\
$\quad$ \\
\textbf{Instructions:} \ldots \hlyellow{
 If the context suggests multiple valid answers, choose the one that best matches the question's wording and the most direct information provided.

 In cases of ambiguity or multiple interpretations, provide an answer that adheres to the most common understanding of the question's terms and context.

} \ldots\\
\textbf{Question:} What is one name for the Cassava after it is marketed as a paste and often served with palm nut soup?
};
\node[labelbox, anchor=south, above=0mm of test-task-2.north] (test-label-2) {\textbf{Inference with meta-reflection}};

\node[box, anchor=north, below=1mm of test-task-2.south] (test-trajectory-2) {
\ldots\\
\textbf{Thought:} The question is asking for a name of cassava when it is marketed as a paste and often served with palm nut soup. From the context provided, I can see that cassava paste is referred to as ``Akpu" in Nigeria and is often eaten with Egusi soup. However, since the question specifically mentions palm nut soup, I should look for a name that is associated with cassava paste served with palm nut soup. The context also mentions ``Plakali" as a staple food made of cassava dough and eaten with palm nut soup in Ghana. However, ``Plakali" is described as cassava dough cooked in hot water, not as a marketed paste. The most direct match for cassava marketed as a paste and associated with palm nut soup is ``Fufu," which is mentioned as a staple food often served with palm nut soup.\\
\hlgreen{\textbf{Action:} Finish[Fufu]}
};

\draw[dotted]
    ($(test-task.north west) + (-1cm, 0.3cm)$)
    to
    node[labelbox, above, rotate=90] {Offline phase}
    node[labelbox, below, rotate=90] {Online phase}
    ($(test-trajectory-2.south west) + (-1cm, 0cm)$);

\draw[double]
    ($(test-trajectory.south west) + (-0.3cm,-0.3cm)$)
    to
    ($(test-trajectory.south east) + (+0.3cm,-0.3cm)$);

\draw[->] (instructions.east) to[out=0, in=180, looseness=0.7] (test-task-2.west) ;

\end{tikzpicture}
\caption{\MetaReflection in the \hotpot domain with  \COT style of reasoning}
\label{fig:CoT-hotpotqa}
\end{figure*}
}

\end{document}